\documentclass[twoside]{article}

\usepackage[accepted]{aistats2024}

\usepackage[round]{natbib}

\usepackage[pdftex]{color}
\definecolor{blueviolet}{cmyk}{0.86,0.91,0.00,0.04}
\usepackage[hidelinks,pdftex]{hyperref}

\hypersetup{
    colorlinks=true,
    citecolor=blueviolet,
    linkcolor=blueviolet,
    urlcolor=blueviolet
}

\usepackage{amsmath,amssymb,amsfonts,amsthm}
\usepackage{booktabs}
\usepackage{bm}
\usepackage{dsfont}
\usepackage{subcaption}
\usepackage{url}
\usepackage{here}
\usepackage{multirow}
\usepackage{mathtools}

\usepackage[linesnumbered,ruled]{algorithm2e}
\SetKwInput{KwInput}{Input}
\SetKwInput{KwOutput}{Output}
\SetAlFnt{\small}
\SetAlCapFnt{\small}
\SetAlCapNameFnt{\small}

\newcommand{\argmax}{\mathop{\rm arg~max}\limits}

\theoremstyle{definition}
\newtheorem*{defi*}{Definition}
\newtheorem{rem}{Remark}

\theoremstyle{plain}
\newtheorem{theo}{Theorem}
\newtheorem{cor}{Corollary}

\allowdisplaybreaks[4]

\begin{document}

\runningauthor{Yasushi Esaki, Akihiro Nakamura, Keisuke Kawano, Ryoko Tokuhisa, Takuro Kutsuna}

\twocolumn[
\aistatstitle{Accuracy-Preserving Calibration via Statistical Modeling on Probability Simplex}
\aistatsauthor{ Yasushi Esaki$^\dag$$^*$ \And Akihiro Nakamura$^\dag$ \And  Keisuke Kawano$^\dag$ }
\vspace{0.5cm}
\aistatsauthor{ Ryoko Tokuhisa$^\dag$ \And  Takuro Kutsuna$^\dag$ }
\vspace{0.5cm}
\aistatsaddress{ $^\dag$Toyota Central R\&D Labs., Inc.
}]

\begin{abstract}
   Classification models based on deep neural networks (DNNs) must be calibrated to measure the reliability of predictions. Some recent calibration methods have employed a probabilistic model on the probability simplex. However, these calibration methods cannot preserve the accuracy of pre-trained models, even those with a high classification accuracy. We propose an accuracy-preserving calibration method using the Concrete distribution as the probabilistic model on the probability simplex. We theoretically prove that a DNN model trained on cross-entropy loss has optimality as the parameter of the Concrete distribution. We also propose an efficient method that synthetically generates samples for training probabilistic models on the probability simplex. We demonstrate that the proposed method can outperform previous methods in accuracy-preserving calibration tasks using benchmarks. The code is available at \url{https://github.com/ToyotaCRDL/SimplexTS}.
\end{abstract}

\section{INTRODUCTION}
\label{sec:intro}
To ensure the safety of systems operated by deep neural network (DNN) classification models, the reliability of the model predictions must be measured for each input.
If the confidence~\citep{calibration} computed by the models matches the classification accuracy, we can confidently measure the reliability of the predictions for unlabeled inputs.
Many previous methods train the models until the confidence matches the accuracy~\citep{self-training,generalization}. Such a training scheme is called calibration~\citep{calibration}. However, models trained under calibration tend to have lower accuracy than models trained only for classification accuracy~\citep{mixup-calibration}. Degrading the accuracy to improve the model calibration is undesired. In this paper, we instead aim to calibrate the confidence of a highly accurate model trained without considering calibration while preserving its accuracy. This approach is called accuracy-preserving calibration~\citep{ETS}.

While many previous researches on calibration have adjusted the input-dependent parameter of the categorical distribution~\citep{calibration-survay}, some existing studies optimize the input-dependent parameter of the Dirichlet distribution for calibration~\citep{EnD2,ScalingEnD2}. A probabilistic model on the probability simplex, such as the Dirichlet distribution, enables us to distinguish between two types of prediction uncertainty, aleatoric uncertainty and epistemic uncertainty~\citep{uncertainty}, which cannot be distinguished by the categorical distribution. This advantage prevents overconfidence, which is caused by failure to estimate the epistemic uncertainty~\citep{overconfidence}.
However, the above methods~\citep{EnD2,ScalingEnD2} cannot preserve the model accuracy because a DNN-based classification model is updated for optimizing the input-dependent parameter of the Dirichlet distribution.
In addition, these methods obtain labels on the probability simplex from ensemble models to estimate the Dirichlet distribution. Since ensemble models comprise training a few tens of DNN models initialized with different random weight parameters, the training overhead is excessively large.

The existing accuracy-preserving calibration methods, such as Temperature Scaling (TS)~\citep{calibration}, introduce a temperature parameter separately from the logits of a pre-trained DNN model for the confidence computation. The temperature parameter is optimized to improve confidence while freezing the pre-trained DNN model.
However, TS only adjusts the parameter of the categorical distribution and does not optimize the parameter of the probabilistic model on the probability simplex.
Although many extensions to the original TS method~\citep{ETS,PTS,AdaptiveTemp} have been proposed, these methods also adjust the parameter of the categorical distribution.
Therefore, the previous TS methods have difficulty distinguishing between the aleatoric uncertainty and the epistemic uncertainty, which limits the calibration performance.

Therefore, an accuracy-preserving calibration method that estimates a probabilistic model on the probability simplex is required. 
One possible approach is to combine TS and the Dirichlet distribution.\footnote{An example is presented in Appendix~\ref{sec:dirichlet}.}
However, this approach has problems in optimizing the temperature parameter alone (see Section~\ref{sec:location}) and shows low calibration performance in our experiments (see Section~\ref{sec:dirnew}).

In this paper, we propose Simplex TS (STS), which applies the Concrete distribution~\citep{concrete} as the probabilistic model on the probability simplex.
The Concrete distribution has two parameters, the location parameter and the temperature parameter.
STS optimizes these two parameters in turn, corresponding to the training for classification and the subsequent accuracy-preserving calibration, respectively.
More precisely, we prove theoretically that a DNN model trained with cross-entropy loss optimizes the location parameter independently of the temperature parameter.
This fact allows us to reuse the pre-trained DNN model for optimizing the location parameter and optimize the temperature parameter solely for calibration.
Both parameters of the Concrete distribution are optimized by the accuracy-preserving calibration alone, if we already have the pre-trained DNN model.
To optimize the temperature parameter, we also propose Multi-Mixup, which synthetically generates training samples labeled with vectors on the probability simplex. Multi-Mixup does not have to train ensemble models and ameliorates the training overhead.
Through numerical experiments, we demonstrate that STS can outperform previous TS methods in accuracy-preserving calibration tasks.

Our contributions are summarized below.
\begin{itemize}
    \item We propose an accuracy-preserving calibration method using the Concrete distribution as a probabilistic model on the probability simplex (Section~\ref{sec:method}) and demonstrate that the proposed method outperforms previous methods (Section~\ref{sec:ex}). 
    \item We propose a method that synthetically generates a dataset for training probabilistic models on the probability simplex. This method reduces the training overhead from those of the previous ensemble methods (Section~\ref{sec:multi-mixup}).
\end{itemize}

\section{RELATED WORK}
\label{sec:related} 
\subsection{Accuracy-Preserving Calibration}
\label{sec:post-hoc}
TS~\citep{calibration} is a typical accuracy-preserving calibration method. 
As mentioned above, researchers have proposed many successor methods ~\citep{ETS,AdaptiveTemp2} that extend the original TS method~\citep{calibration}.
Some recent TS methods attempt to optimize the temperature parameter for each sample with a function that depends on the inputs~\citep{PTS,AdaptiveTemp}.
The commonality of these TS methods is that they improve confidence by tuning a scalar-valued temperature parameter separately from a pre-trained DNN model.
Accuracy-preserving calibration methods other than TS are also available~\citep{DensityEstimation,non-parametric,DS}. \citet{DensityEstimation} train an auxiliary model to transform the softmax outputs of a pre-trained model for calibration and \citet{non-parametric} optimize the parameter of the categorical distribution following the Gaussian process. These studies do not consider a probabilistic model on the probability simplex. \citet{DS} aim to perform out-of-distribution detection while preserving a pre-trained DNN model, which deviates from the calibration in terms of problem settings.
\subsection{Uncertainty Estimation on the Probability Simplex}
\label{sec:unc-calib}
Many studies estimate the prediction uncertainty, which consists of aleatoric uncertainty and epistemic uncertainty~\citep{uncertainty}, using the Dirichlet distribution whose parameter is computed by an input-dependent DNN model~\citep{PriorNet,EDL}.
Prior Networks~\citep{PriorNet} are trained by minimizing the Kullback--Leibler divergence of the Dirichlet distribution.
Evidential Deep Learning~\citep{EDL} measures the amount of evidence of each class by estimating the Dirichlet distribution.
Uncertainty estimation methods that use a probabilistic model on the probability simplex, such as the Dirichlet distribution, can estimate aleatoric and epistemic uncertainties distinctly.
Note that the above studies do not discuss accuracy-preserving calibration, and to the best of our knowledge, no studies have applied a probabilistic model on the probability simplex to the accuracy-preserving calibration. 

\subsection{Data Augmentation by Interpolation}
\label{sec:dataaug}
Some data augmentation methods generate pseudo-new samples by linearly interpolating the original samples~\citep{mixup,mixup-calibration}. Whereas many studies interpolate two samples~\citep{PairingSamples,ManifoldMixup}, AdaMixup~\citep{AdaMixUp} and $\zeta$-Mixup~\citep{zeta-mixup} interpolate three or more samples, which is the same as Multi-Mixup.
AdaMixup and $\zeta$-Mixup aim to improve the classification performance and have mechanisms to restrict the generation of samples that differ significantly from the original samples. This property is not suitable for calibration because the features of inputs with a large prediction uncertainty cannot be learned.

\section{PRELIMINARIES}
\subsection{Notations}
We write a bold symbol $\bm{a}$ for the real vector, and let $a_i$ be the $i$-th entry of~$\bm{a}$ with $i\in\mathbb{N}$. The real vector-valued function is also written as a bold symbol~$\bm{f}$, and let~$f_i(\bm{x})$ be the $i$-th coordinate of the output of~$\bm{f}(\bm{x})$. Furthermore, let $\mathds{1}_{\{\cdot \}}$ be an indicator function and $\Delta_{q-1}$ be the $q-1$ dimensional probability simplex, where $q\in\mathbb{N}$ is an integer. Specifically, $\Delta_{q-1}:=\{\bm{\pi}\in [0,1]^q|\sum^q_{i=1}\pi_i=1\}$. Let $\mathcal{X}(\subseteq\mathbb{R}^d)$ be an input space and let $\mathcal{Y}=\{1,\ldots,K\}$ be a set of class categories, where $K\in\mathbb{N}$ is the number of classes. Let~$\mathrm{Cat}(y|\bm{\pi})$ be the probability mass function of the categorical distribution with a parameter $\bm{\pi}\in\Delta_{K-1}$. Let~$\mathrm{Dir}(\bm{\pi}|\bm{\mu})$ be the probability density function of the Dirichlet distribution with a parameter $\bm{\mu}\in(0, \infty)^q$.
\subsection{Uncertainty Estimation on the Probability Simplex}
\label{sec:uncertainty}
\citet{PriorNet} propose to estimate the prediction uncertainty in classification through a probabilistic model on the probability simplex. Let the class label $y\in\mathcal{Y}$ be a stochastic variable following a probability distribution $p(y|\bm{\pi})$, where~$\bm{\pi}\in\Delta_{K-1}$ is a parameter. Let~$\bm{\pi}$ be another stochastic variable following the conditional probability distribution~$p(\bm{\pi}|\bm{x})$, where $\bm{x}\in\mathcal{X}$ is the input. A conditional class distribution~$p(y|\bm{x})$ given an input $\bm{x}$ can be described as
\begin{align}
p(y|\bm{x})=\int p(y|\bm{\pi})p(\bm{\pi}|\bm{x})d\bm{\pi}. \label{eq:predictive}
\end{align}
To estimate the prediction uncertainty, \citet{PriorNet} estimate $p(\bm{\pi}|\bm{x})$ following the Dirichlet distribution. Let $\bm{\mu}(\bm{x}, \bm{\theta}_{\mathrm{\mu}})=(e^{g_1(\bm{x})},\ldots,e^{g_K(\bm{x})})^\top$, where $\bm{g} : \mathcal{X}\to\mathbb{R}^K$ is a DNN model (a model with logit outputs~$\bm{g}(\bm{x})$) parameterized by a real vector~$\bm{\theta}_{\mathrm{\mu}}$. \citet{PriorNet} assume that~$y$ follows the categorical distribution with a parameter~$\bm{\pi}$, whereas~$\bm{\pi}$ follows the Dirichlet distribution with a $\bm{x}$-dependent parameter~$\bm{\mu}(\bm{x}, \bm{\theta}_{\mathrm{\mu}})$:
\begin{align}
\begin{split}
    p(y|\bm{\pi}) &= \mathrm{Cat}(y|\bm{\pi}),\\
    p(\bm{\pi}|\bm{x}) &= \mathrm{Dir}(\bm{\pi}|\bm{\mu}(\bm{x}, \bm{\theta}_{\mathrm{\mu}})).
\end{split} \label{eq:dir}
\end{align}
If the estimated value of $\bm{\theta}_{\mathrm{\mu}}$ is denoted by~$\hat{\bm{\theta}}_{\mathrm{\mu}}$, the prediction uncertainty is represented by the estimated distribution $\mathrm{Dir}(\bm{\pi}|\bm{\mu}(\bm{x}, \hat{\bm{\theta}}_{\mathrm{\mu}}))$, and we can estimate the prediction uncertainty for unlabeled inputs. The prediction uncertainty is estimated to be small when $\mathrm{Dir}(\bm{\pi}|\bm{\mu}(\bm{x}, \hat{\bm{\theta}}_{\mathrm{\mu}}))$ has a high probability density at one of the apexes of $\Delta_{K-1}$. Conversely, the prediction uncertainty is estimated to be large when~$\mathrm{Dir}(\bm{\pi}|\bm{\mu}(\bm{x}, \hat{\bm{\theta}}_{\mathrm{\mu}}))$ has a high probability density at the center of $\Delta_{K-1}$ or has a large variance over~$\Delta_{K-1}$. Specifically, a high probability density at the center of $\Delta_{K-1}$ corresponds to high aleatoric uncertainty, and a large variance over $\Delta_{K-1}$ corresponds to high epistemic uncertainty~\citep{PriorNet}. Applying a probabilistic model on $\Delta_{K-1}$ such as $\mathrm{Dir}(\bm{\pi}|\bm{\mu}(\bm{x}, \bm{\theta}_{\mu}))$ provides the distinction between aleatoric uncertainty and epistemic uncertainty, which cannot be distinguished when considering only the categorical distribution. This distinction helps confidence take a lower value when the number of training samples is small, corresponding to epistemic uncertainty, and avoids overconfidence~\citep{overconfidence}.

The previous studies discussing uncertainty estimation~\citep{EnD2,generalEnD2,ScalingEnD2} assume that the distribution of the softmax outputs of ensemble models input with~$\bm{x}$ indicates the prediction uncertainty of $\bm{x}$, and estimate~$\bm{\theta}_{\mathrm{\mu}}$ on a dataset labeled with these softmax outputs to learn their distribution.
The Dirichlet distribution is appropriate for learning the prediction uncertainty through the ensemble models because the softmax outputs of the ensemble models are vectors on~$\Delta_{K-1}$ and their distribution is represented by a probabilistic model on~$\Delta_{K-1}$.
Although these studies achieve high calibration performance~\citep{EnD2,generalEnD2,ScalingEnD2}, they have difficulty in performing accuracy-preserving calibration. 
\subsection{Concrete Distribution}
\label{sec:concrete}
The Concrete distribution~\citep{concrete} is a family of probability distributions on the probability simplex. It is formulated by extending the categorical distribution to a continuous probability distribution and is parameterized by two parameters: the location parameter~$\bm{\alpha}\in(0, \infty)^q$ and the temperature parameter~$\lambda\in(0, \infty)$. The probability density function of the Concrete distribution is defined as follows.
\begin{defi*}[Concrete distribution; \citep{concrete}]\label{def:concrete}
Let $\bm{\alpha}\in(0, \infty)^q$ and $\lambda\in(0, \infty)$. The probability density of a stochastic variable $\bm{\pi}\in\Delta_{q-1}$ following the Concrete distribution with location parameter~$\bm{\alpha}$ and temperature parameter $\lambda$ is given by
\begin{align}
    \mathrm{Cn}(\bm{\pi}|\bm{\alpha}, \lambda) \coloneqq (q-1)!\lambda^{q-1}\prod^q_{j=1}\left(\frac{\alpha_j\pi^{-(\lambda+1)}_j}{\sum^q_{i=1}\alpha_i\pi^{-\lambda}_i}\right). \label{eq:density}
\end{align}
\end{defi*}
The location parameter $\bm{\alpha}$ indicates the relative mass among the apexes of $\Delta_{q-1}$, and the temperature parameter $\lambda$ indicates the difference from the categorical distribution whose parameter is given by~$(\alpha_1/\sum^{q}_{i=1}\alpha_i,\ldots,\alpha_q/\sum^{q}_{i=1}\alpha_i)^\top$. The Concrete distribution has been adopted as the reparameterization trick in a variational autoencoder~\citep{concrete,gumbelsoft}. To the best of our knowledge, no research has used this distribution for calibration purposes.
\begin{rem}[Properties of the temperature parameter]\label{em1}
The variance over $\Delta_{K-1}$ can be changed by shifting $\lambda$. Moreover, the probability density is concentrated at the apexes of $\Delta_{q-1}$ if $\lambda$ is sufficiently small and at the center of $\Delta_{q-1}$ if $\lambda$ is sufficiently large~\citep{concrete}. Note that this variation is realized even when $\bm{\alpha}$ is fixed. This property of~$\lambda$ implies that optimizing $\lambda$ on a dataset that has labels indicating the prediction uncertainty contributes to uncertainty estimation.
\end{rem}
\section{SIMPLEX TEMPERATURE SCALING}
\label{sec:method}
In this section, we propose STS for accuracy-preserving calibration. We first present the overview of STS in Section~\ref{sec:prop}. We then show theoretically that STS enables accuracy-preserving calibration in Section~\ref{sec:location}. 
The training of STS with the data generation method is described in Section~\ref{sec:temperature}.
Section~\ref{sec:confidence} covers the confidence calculation, while the implementation of STS is shown in Section~\ref{sec:archit}.
\subsection{Overview}
\label{sec:prop}
This subsection introduces a probabilistic model and its optimization process for STS.

\paragraph{Probabilistic model.}
Let~$\bm{\alpha}(\cdot, \bm{\theta}_{\mathrm{\alpha}}) : \mathcal{X}\to(0, \infty)^K$ and $\lambda(\cdot, \bm{\theta}_{\mathrm{\lambda}}) : \mathcal{X}\to(0, \infty)$ be deterministic functions parameterized by real vectors $\bm{\theta}_{\mathrm{\alpha}}$ and $\bm{\theta}_{\mathrm{\lambda}}$, respectively. Our probabilistic models for $p(y|\bm{\pi})$ and $p(\bm{\pi}|\bm{x})$ are respectively formulated as
\begin{align}
    p(y=k|\bm{\pi})&=\mathds{1}_{\{\mathrm{arg}\max_{i\in\mathcal{Y}}\pi_i=k\}}\ \ \ \mbox{for}\ \forall{k}\in\mathcal{Y}, \label{eq:ind}\\
    p(\bm{\pi}|\bm{x})&=\mathrm{Cn}(\bm{\pi}|\bm{\alpha}(\bm{x}, \bm{\theta}_{\mathrm{\alpha}}), \lambda(\bm{x}, \bm{\theta}_{\mathrm{\lambda}})). \label{eq:con}
\end{align}
As indicated in Eq.~\eqref{eq:ind}, $p(y|\bm{\pi})$ is given by a one-hot vector in which one component is~1 and the other components are~0. That is, the value of~$y$ is definitively determined from $\bm{\pi}$, meaning that $y$ is a stochastic variable if and only if $\bm{\pi}$ is a stochastic variable.
Although Eq.~\eqref{eq:ind} has no variance, the combination of Eqs.~\eqref{eq:ind} and \eqref{eq:con} has good properties for consistently formulating the pre-training for classification and the subsequent accuracy-preserving calibration. 

\paragraph{Optimization process.}
The pre-training for classification and the subsequent accuracy-preserving calibration can be formulated as consecutively estimating~$\bm{\theta}_{\mathrm{\alpha}}$ and $\bm{\theta}_{\mathrm{\lambda}}$ using different criteria. Given a dataset $\mathcal{D}\coloneqq\{(\bm{x}_n, y_n)\}^N_{n=1}$, where $y_n\in\mathcal{Y}$ is the class label of $\bm{x}_n\in\mathcal{X}$, STS performs the following two steps.
\begin{description}
    \item[Step 1: Classification] Based on $\mathcal{D}$, we estimate $\bm{\theta}_{\mathrm{\alpha}}$ via maximum likelihood estimation of Eq.~\eqref{eq:predictive} under Eqs.~\eqref{eq:ind} and~\eqref{eq:con}. The estimated value of $\bm{\theta}_{\mathrm{\alpha}}$ is denoted by~$\hat{\bm{\theta}}_{\mathrm{\alpha}}$.
    \item[Step 2: Calibration] Based on $\tilde{\mathcal{D}}$, we estimate $\bm{\theta}_{\mathrm{\lambda}}$ via maximum likelihood estimation of Eq.~\eqref{eq:con}, fixing~$\bm{\theta}_{\mathrm{\alpha}}$ to~$\hat{\bm{\theta}}_{\mathrm{\alpha}}$. The estimated value of~$\bm{\theta}_{\mathrm{\lambda}}$ is denoted by $\hat{\bm{\theta}}_{\mathrm{\lambda}}$.
\end{description}
In Step~2, $\tilde{\mathcal{D}}(\coloneqq \{(\tilde{\bm{x}}_m, \tilde{\bm{\pi}}_m)\}^M_{m=1})$ is a dataset in which the inputs $\tilde{\bm{x}}_1,\ldots,\tilde{\bm{x}}_M\in\mathcal{X}$ and their prediction uncertainties $\tilde{\bm{\pi}}_1,\ldots,\tilde{\bm{\pi}}_M\in\Delta_{K-1}$ have been synthetically generated from original samples with hard labels. Since original classification datasets have only hard labels corresponding to the apexes of $\Delta_{K-1}$, we obtain labels on $\Delta_{K-1}$ by synthetically generating $\tilde{\mathcal{D}}$. Note that in practice we do not need to pre-determine~$M$ and to generate the whole of~$\tilde{\mathcal{D}}$ before Step~2, since we can generate as many samples in~$\tilde{\mathcal{D}}$ as we need during the optimization in Step~2. The details of generating samples in $\tilde{\mathcal{D}}$ are explained in Section~\ref{sec:multi-mixup}.

In the following subsection, we show the following properties of the proposed model:
1)~Given a classifier trained with ordinary cross-entropy loss on~$\mathcal{D}$, the parameter $\hat{\bm{\theta}}_{\mathrm{\alpha}}$, which is estimated in Step~1, is immediately obtained. 2)~The~$\hat{\bm{\theta}}_{\mathrm{\alpha}}$ is optimal regardless of~$\bm{\theta}_{\mathrm{\lambda}}$. 
3)~Step~2 preserves the classification accuracy of the classifier in Step~1.  
These properties indicate that we can calibrate a given pre-trained classifier without changing its classification accuracy, where~$\bm{\theta}_{\mathrm{\lambda}}$ is introduced as a new parameter for better calibration.
\subsection{Why STS Realizes Accuracy-Preserving Calibration}
\label{sec:location}
In this subsection, we present the reason why the probabilistic model and the optimization process introduced in Section~\ref{sec:prop} serve as classification and accuracy-preserving calibration.
To show that Step~1 corresponds to classification, we discuss Eq.~\eqref{eq:predictive} under Eqs.~\eqref{eq:ind} and~\eqref{eq:con}.
Let $\bm{\alpha}(\bm{x}, \bm{\theta}_{\mathrm{\alpha}})=(e^{g_1(\bm{x})},\ldots,e^{g_K(\bm{x})})^\top$, where $\bm{g} : \mathcal{X}\to\mathbb{R}^K$ is a DNN model parameterized by~$\bm{\theta}_{\mathrm{\alpha}}$. In this case, Eq.~\eqref{eq:predictive} becomes the softmax function with~$\bm{g}(\bm{x})$ as logits, from the following theorem. 
\begin{theo}[Predictive distribution]\label{prop2}
We assume that $p(y|\bm{\pi})$ and $p(\bm{\pi}|\bm{x})$ are formulated as Eqs.~\eqref{eq:ind} and~\eqref{eq:con}. 
Then, Eq.~\eqref{eq:predictive} is given as follows.\footnote{For confidence, another criterion is applied instead of Eq.~\eqref{eq:predictive}. The confidence is explained in Section~\ref{sec:confidence}.}
\begin{align}
\begin{split}
    p(y=k|\bm{x})&=\int p(y=k|\bm{\pi})p(\bm{\pi}|\bm{x})d\bm{\pi}\\
        &=\frac{\alpha_k(\bm{x}, \bm{\theta}_{\mathrm{\alpha}})}{\sum^K_{i=1}\alpha_i(\bm{x}, \bm{\theta}_{\mathrm{\alpha}})}\ \ \ \mbox{for}\ \forall{k}\in\mathcal{Y}. \label{eq:previous2}
\end{split}
\end{align}
\end{theo}
The proof is shown in~Appendix~\ref{sec:proof2}. Note that Eq.~\eqref{eq:previous2} is independent of $\lambda(\bm{x}, \bm{\theta}_{\mathrm{\lambda}})$. From Theorem~\ref{prop2}, we can show that Step~1 is equivalent to the standard DNN training with the cross-entropy loss as follows.
\begin{cor}[Classification via the Concrete distribution]\label{prop1}
Let $\mathcal{D}=\{(\bm{x}_n, y_n)\}^N_{n=1}$ and~$\bm{\alpha}(\bm{x}, \bm{\theta}_{\mathrm{\alpha}})=(e^{g_1(\bm{x})},\ldots,e^{g_K(\bm{x})})^\top$, where $\bm{g} : \mathcal{X}\to\mathbb{R}^K$ is a function parameterized by~$\bm{\theta}_{\mathrm{\alpha}}$. 
Under Eqs.~\eqref{eq:ind} and~\eqref{eq:con}, optimizing~$\bm{\theta}_{\mathrm{\alpha}}$ on~$\mathcal{D}$ using the maximum likelihood estimation of Eq.~\eqref{eq:predictive} is equivalent to optimizing~$\bm{\theta}_{\mathrm{\alpha}}$ on~$\mathcal{D}$ using the minimization of the following criterion.
\begin{align}
    \mathcal{L}(\bm{\theta}_{\mathrm{\alpha}})=-\frac{1}{N}\sum^N_{n=1}\sum^K_{k=1}\mathds{1}_{\{y_n=k\}}\ln \frac{e^{g_k(\bm{x}_n)}}{\sum^K_{i=1}e^{g_i(\bm{x}_n)}}. \label{eq:cross}
\end{align}
\end{cor}
\begin{rem}[Interpretations of Corollary~\ref{prop1}]\label{rem:int}
Corollary~\ref{prop1} shows that the weight parameters of a DNN model trained with the cross-entropy loss can be regarded as~$\hat{\bm{\theta}}_{\mathrm{\alpha}}$, which is estimated in Step~1.
If the pre-trained DNN model is denoted by $\hat{\bm{g}} : \mathcal{X}\to\mathbb{R}^K$, we have $\bm{\alpha}(\bm{x}, \hat{\bm{\theta}}_{\mathrm{\alpha}})=(e^{\hat{g}_1(\bm{x})},\ldots,e^{\hat{g}_K(\bm{x})})^\top$.
Therefore, we can reuse the pre-trained DNN model for~$\bm{\alpha}(\bm{x}, \hat{\bm{\theta}}_{\mathrm{\alpha}})$ and skip Step~1. The proof is shown in Appendix~\ref{sec:proof1}.
\end{rem}

The maximality of the likelihood of Eq.~\eqref{eq:predictive} (i.e., the minimality of the cross-entropy loss) is guaranteed if~$\bm{\theta}_{\mathrm{\lambda}}$ shifts, because Eq.~\eqref{eq:previous2} is independent of $\lambda(\bm{x}, \bm{\theta}_{\mathrm{\lambda}})$.
This fact motivates us to fix $\bm{\alpha}(\bm{x}, \hat{\bm{\theta}}_{\mathrm{\alpha}})$ in Step~2.
As mentioned in Remark~\ref{em1}, we can estimate the prediction uncertainty by optimizing $\lambda(\bm{x}, \bm{\theta}_{\mathrm{\lambda}})$ on $\tilde{\mathcal{D}}$, even when $\bm{\alpha}(\bm{x}, \hat{\bm{\theta}}_{\mathrm{\alpha}})$ is fixed. Therefore, Step~2 realizes a calibration based on the prediction uncertainty.

\paragraph{Preserved accuracy.}
If we consider predicting the class using Eq.~\eqref{eq:predictive} after Step~2, the classification accuracy of the pre-trained DNN model is preserved.
According to~Theorem~\ref{prop2}, if we have~$\hat{\bm{\theta}}_{\mathrm{\alpha}}$, the predicted class is computed as
\begin{align}
    \argmax_{k\in\mathcal{Y}}p(y=k|\bm{x})=\argmax_{k\in\mathcal{Y}}\frac{\alpha_k(\bm{x}, \hat{\bm{\theta}}_{\mathrm{\alpha}})}{\sum^K_{i=1}\alpha_i(\bm{x}, \hat{\bm{\theta}}_{\mathrm{\alpha}})}. \label{eq:ps}
\end{align}
Equation~\eqref{eq:ps} is equal to the class predicted by the pre-trained DNN model $\hat{\bm{g}}$ because $\bm{\alpha}(\bm{x}, \hat{\bm{\theta}}_{\mathrm{\alpha}})=(e^{\hat{g}_1(\bm{x})},\ldots,e^{\hat{g}_K(\bm{x})})^\top$ as in Remark~\ref{rem:int}. Moreover, this prediction does not change when $\bm{\theta}_{\mathrm{\lambda}}$ is updated in Step~2.
As a result, the classification accuracy of the pre-trained DNN model is preserved. 

\paragraph{Problems with the Dirichlet distribution.}
The probabilistic model with the Dirichlet distribution as in Eq.~\eqref{eq:dir} also leads to an equation similar to Eq.~\eqref{eq:previous2} such as
\begin{align}
\begin{split}
    p(y=k|\bm{x})&=\int p(y=k|\bm{\pi})p(\bm{\pi}|\bm{x})d\bm{\pi}\\
        &=\frac{\mu_k(\bm{x}, \bm{\theta}_{\mathrm{\mu}})}{\sum^K_{i=1}\mu_i(\bm{x}, \bm{\theta}_{\mathrm{\mu}})}\ \ \ \mbox{for}\ \forall{k}\in\mathcal{Y}.
\end{split}\label{eq:pt}
\end{align}
If $\bm{\mu}(\bm{x}, \bm{\theta}_{\mathrm{\mu}})=(e^{g_1(\bm{x})},\ldots,e^{g_K(\bm{x})})^\top$, where $\bm{g}$ is parameterized by $\bm{\theta}_{\mathrm{\mu}}$, Eq.~\eqref{eq:pt} becomes the softmax function. Therefore, the maximum likelihood estimation of Eq.~\eqref{eq:predictive} is equivalent to minimizing the cross-entropy loss, as in STS.
However, the Dirichlet distribution does not have a temperature parameter independent of $\bm{\mu}(\bm{x}, \bm{\theta}_{\mathrm{\mu}})$.
Even if we introduce a temperature parameter in $\bm{\mu}(\bm{x}, \bm{\theta}_{\mathrm{\mu}})$, the maximality of the likelihood of Eq.~\eqref{eq:predictive} (i.e., the minimality of the cross-entropy loss) is not guaranteed after optimizing the temperature parameter solely using the maximum likelihood estimation of $\mathrm{Dir}(\bm{\pi}|\bm{\mu}(\bm{x}, \bm{\theta}_{\mathrm{\mu}}))$.\footnote{An example is presented in Appendix~\ref{sec:dirichlet}.}
Our STS solves this problem by replacing the Dirichlet distribution with the Concrete distribution and enables us to perform an accuracy-preserving calibration.
\subsection{Training of the Temperature Parameter}
\label{sec:temperature}
This subsection describes the dataset and objective function of Step 2.
\subsubsection{Generation of Samples on Probability Simplex by Multi-Mixup}
\label{sec:multi-mixup}
To estimate $\bm{\theta}_{\mathrm{\lambda}}$ by the maximum likelihood process, we require a dataset $\tilde{\mathcal{D}}$ in which inputs are labeled with vectors on the probability simplex that indicate the prediction uncertainty.
As the model must distinguish inputs with large prediction uncertainties from those with small prediction uncertainties, the dataset should include both input types.
As in explained Section~\ref{sec:uncertainty}, previous calibration methods using Eq.~\eqref{eq:dir} train ensemble models and label the inputs with the softmax outputs of the trained ensemble models, thus obtaining a dataset that satisfies the above properties~\citep{EnD2,generalEnD2,ScalingEnD2}. Composing a dataset through this process requires training a few tens of DNN models initialized with different random weight parameters, which incurs a large training overhead. In addition, ensemble models that follow a pre-trained DNN model are difficult to obtain when the pre-trained DNN model is published online.

To solve these problems, we propose a method that synthetically generates training samples with labels on the probability simplex. Our Multi-Mixup method randomly selects one sample from each class $k\in\mathcal{Y}$ and linearly interpolates the inputs and one-hot labels of the selected samples.
Let $\bm{y}^{(1)},\ldots,\bm{y}^{(K)}$ be one-hot vectors indicating the classes. The $k$-th element of $\bm{y}^{(k)}$ is 1 and all other elements are 0. Now let~$\bm{x}^{(1)},\ldots,\bm{x}^{(K)}$ be the inputs in $\mathcal{X}$ corresponding to $\bm{y}^{(1)},\ldots,\bm{y}^{(K)}$. Multi-Mixup performs the following interpolation:
\begin{align}
    \tilde{\bm{x}}=\sum^K_{k=1}w_k\bm{x}^{(k)}, \ \ \ \ \tilde{\bm{\pi}}=\sum^K_{k=1}w_k\bm{y}^{(k)}, \label{eq:mix}
\end{align}
where $\bm{w}$ is a random variable on $\Delta_{K-1}$. The details of Multi-Mixup, including how to determine the value of $\bm{w}$, are explained in Appendix~\ref{sec:alg}.

Multi-Mixup can automatically generate inputs with small or large prediction uncertainties, along with labels indicating their prediction uncertainties. For example, we consider two weights, $\bm{w}=(0.01, 0.01, 0.98)^\top$ and $\bm{w}=(\frac{1}{3}, \frac{1}{3}, \frac{1}{3})^\top$, under $K=3$. The first weight generates an input with a small prediction uncertainty near the input~$\bm{x}^{(3)}$, namely, $\tilde{\bm{x}}=0.01\bm{x}^{(1)}+0.01\bm{x}^{(2)}+0.98\bm{x}^{(3)}$. The second weight generates an input with a large prediction uncertainty far from any of the three inputs $\bm{x}^{(1)}$, $\bm{x}^{(2)}$, or $\bm{x}^{(3)}$, which is computed as $\tilde{\bm{x}}=\frac{1}{3}(\bm{x}^{(1)}+\bm{x}^{(2)}+\bm{x}^{(3)})$.
These two inputs are labeled with $\tilde{\bm{\pi}}=(0.01, 0.01, 0.98)^\top$ and $\tilde{\bm{\pi}}=(\frac{1}{3}, \frac{1}{3}, \frac{1}{3})^\top$, respectively. These labels indicate the prediction uncertainty of the two inputs. 

\paragraph{Problems with existing data augmentations.} Mixup~\citep{mixup} generates samples only on the edges of~$\Delta_{K-1}$, which is inadequate for optimizing the parameter of a probabilistic model on~$\Delta_{K-1}$.
Although AdaMixup~\citep{AdaMixUp} and $\zeta$-Mixup~\citep{zeta-mixup} generate samples in the interior of~$\Delta_{K-1}$, the weights of the interpolations are constrained to restricted areas. This constraint is inappropriate for calibration because both inputs with large and small prediction uncertainty should be learned during calibration.
\subsubsection{Loss Function for Training}
\label{sec:loss}
Using the maximum likelihood estimation of Eq.~\eqref{eq:con}, Step 2 of our method estimates $\bm{\theta}_{\mathrm{\lambda}}$ on $\tilde{\mathcal{D}}$ generated by Multi-Mixup.
To estimate the maximum likelihood on~$\tilde{\mathcal{D}}$, we compute the gradient descent~\citep{gradient-descent} of the following negative log-likelihood.
\begin{align}
\begin{split}
      &\mathcal{L}_{\tilde{\mathcal{D}}}(\bm{\theta}_{\mathrm{\lambda}}) = -\frac{1}{M}\sum^M_{m=1} (K-1)\ln \lambda(\tilde{\bm{x}}_m, \bm{\theta}_{\mathrm{\lambda}})                                                                          \\
      &  -\frac{1}{M}\sum^M_{m=1} \sum^K_{k=1}\ln \alpha_k(\tilde{\bm{x}}_m, \hat{\bm{\theta}}_{\mathrm{\alpha}}) \\
      &+\frac{1}{M}\sum^M_{m=1} \sum^K_{k=1}(\lambda(\tilde{\bm{x}}_m, \bm{\theta}_{\mathrm{\lambda}})+1)\ln \tilde{\pi}_{m,k} 
      \\
      &+ \frac{1}{M}\sum^M_{m=1}K\ln \left(\sum^K_{i=1} \alpha_i(\tilde{\bm{x}}_m,\hat{\bm{\theta}}_{\mathrm{\alpha}})\tilde{\pi}^{-\lambda(\tilde{\bm{x}}_m, \bm{\theta}_{\mathrm{\lambda}})}_{m,i}\right),
\end{split} \label{eq:loss}
  \end{align}
where $\tilde{\mathcal{D}}=\{(\tilde{\bm{x}}_m, \tilde{\bm{\pi}}_m)\}^M_{m=1}$ and $\tilde{\pi}_{m,k}$ is the $k$-th entry of~$\tilde{\bm{\pi}}_m$. Equation~\eqref{eq:loss} is derived by taking the negative log of Eq.~\eqref{eq:density}.
\subsection{Confidence}
\label{sec:confidence}
In previous studies, the maximum of the predictive distribution $\max_{k\in\mathcal{Y}}p(y=k|\bm{x})$ is commonly used as confidence~\citep{calibration-survay}. However, in STS, this metric is computed as
\begin{align}
\max_{k\in\mathcal{Y}}p(y=k|\bm{x})&=\max_{k\in\mathcal{Y}}\frac{\alpha_k(\bm{x}, \hat{\bm{\theta}}_{\mathrm{\alpha}})}{\sum^K_{i=1}\alpha_i(\bm{x}, \hat{\bm{\theta}}_{\mathrm{\alpha}})}\\
&=\max_{k\in\mathcal{Y}}\frac{e^{\hat{g}_k(\bm{x})}}{\sum^K_{i=1}e^{\hat{g}_i(\bm{x})}},
\end{align}
because $p(y=k|\bm{x})$ is derived as shown in Theorem~\ref{prop2}. This means that if we use $\max_{k\in\mathcal{Y}}p(y=k|\bm{x})$ as confidence after Step~2, we cannot improve the confidence of the pre-trained DNN model $\hat{\bm{g}}$. Therefore, we propose to use another metric as confidence instead of $\max_{k\in\mathcal{Y}}p(y=k|\bm{x})$.
Let~$p(\bm{\pi}|\bm{x})=\mathrm{Cn}(\bm{\pi}|\bm{\alpha}(\bm{x}, \hat{\bm{\theta}}_{\mathrm{\alpha}}), \lambda(\bm{x}, \hat{\bm{\theta}}_{\mathrm{\lambda}}))$. The proposed confidence is defined as follows.\footnote{The confidence shown in Eq.~\eqref{eq:confidence} measures the sum of the aleatoric uncertainty and the epistemic uncertainty. Appendix~\ref{sec:distinction} describes how to measure the two uncertainties separately after we obtain~$\mathrm{Cn}(\bm{\pi}|\bm{\alpha}(\bm{x}, \hat{\bm{\theta}}_{\mathrm{\alpha}}), \lambda(\bm{x}, \hat{\bm{\theta}}_{\mathrm{\lambda}}))$.}
\begin{align}
\mathrm{confidence}(\bm{x})&\coloneqq \max_{k\in \mathcal{Y}}(\mathbb{E}[\bm{\pi}])_k\\
&=\max_{k\in \mathcal{Y}}\left(\int \bm{\pi}p(\bm{\pi}|\bm{x})d\bm{\pi}\right)_k, \label{eq:confidence}
\end{align}
where $(\cdot)_k$ is the $k$-th entry of the vector in parentheses and $\mathbb{E}[\cdot]$ is the expectation of $p(\bm{\pi}|\bm{x})$. The motivation for using $\max_{k\in \mathcal{Y}}(\mathbb{E}[\bm{\pi}])_k$ as confidence is that in the categorical-Dirichlet model in Eq.~\eqref{eq:dir}, $\max_{k\in \mathcal{Y}}(\mathbb{E}[\bm{\pi}])_k$ becomes equivalent to $\max_{k \in \mathcal{Y}}p(y=k|\bm{x})$, the maximum of Eq.~\eqref{eq:pt}.

The expectation of the Concrete distribution, which cannot be computed analytically, is approximated as the sample mean of the values randomly sampled from~$\mathrm{Cn}(\bm{\pi}|\bm{\alpha}(\bm{x}, \hat{\bm{\theta}}_{\mathrm{\alpha}}), \lambda(\bm{x}, \hat{\bm{\theta}}_{\mathrm{\lambda}}))$.
According to Proposition~1(a) in~\citet{concrete}, the Concrete distribution can be sampled using the standard Gumbel distribution~\citep{Gumbel}. Therefore, the confidences can be computed as follows, where $j=1,\ldots,p$ and $k=1,\ldots,K$.
\begin{align}
    G^{(j)}_k &\sim \mathrm{Gumbel}(0,1),     \label{eqg}                                   \\     
    \hat{\pi}^{(j)}_k(\bm{x})  & =\frac{\exp\left(\frac{\ln \alpha_k(\bm{x}, \hat{\bm{\theta}}_{\mathrm{\alpha}})+G^{(j)}_k}{\lambda(\bm{x}, \hat{\bm{\theta}}_{\mathrm{\lambda}})}\right)}{{\displaystyle \sum^K_{i=1}}\exp\left(\frac{\ln \alpha_i(\bm{x}, \hat{\bm{\theta}}_{\mathrm{\alpha}})+G^{(j)}_i}{\lambda(\bm{x}, \hat{\bm{\theta}}_{\mathrm{\lambda}})}\right)}, \label{eq:pi}\\ 
    \mathrm{confidence}(\bm{x})  &\simeq \max_{k\in\mathcal{Y}}\frac{1}{p}\sum^p_{j=1}\hat{\pi}^{(j)}_k(\bm{x}). \label{eq:expect2}
  \end{align}
  In Eq.~\eqref{eqg}, $\mathrm{Gumbel}(0,1)$ denotes the standard Gumbel distribution.
  This expression states that pseudorandom numbers are generated $p\times K$ times from the standard Gumbel distribution. We set $p=30$ in the experiments of Section~\ref{sec:ex}. The forward computations of~$\bm{\alpha}(\bm{x}, \hat{\bm{\theta}}_{\mathrm{\alpha}})$ and $\lambda(\bm{x}, \hat{\bm{\theta}}_{\mathrm{\lambda}})$ are performed only once when computing the confidence of each input. Therefore, the computational overhead required to calculate confidence is negligible. If confidence is not required, the forward computation of the pre-trained DNN model is sufficient for predicting the class of~$\bm{x}$, negating the need for sampling, as explained in Section~\ref{sec:location}.
  As discussed in Remark~\ref{em1}, the temperature parameter~$\lambda(\bm{x}, \bm{\theta}_{\mathrm{\lambda}})$ can represent the prediction uncertainty of $\bm{x}$, and this parameter is optimized by Multi-Mixup, which reflects the prediction uncertainty of the training samples, in Step~2. Therefore, Eq.~\eqref{eq:expect2} calculated with $\lambda(\bm{x}, \hat{\bm{\theta}}_{\mathrm{\lambda}})$ can approximate the classification accuracy more strictly than the confidence in previous studies. This fact is borne out by the experimental results in Section~\ref{sec:ex}.
\subsection{Architectures for Parameters in Probabilistic Model}
\label{sec:archit}
\begin{figure}[t]
    \centering
      \includegraphics[width=8.7cm, height=6cm]{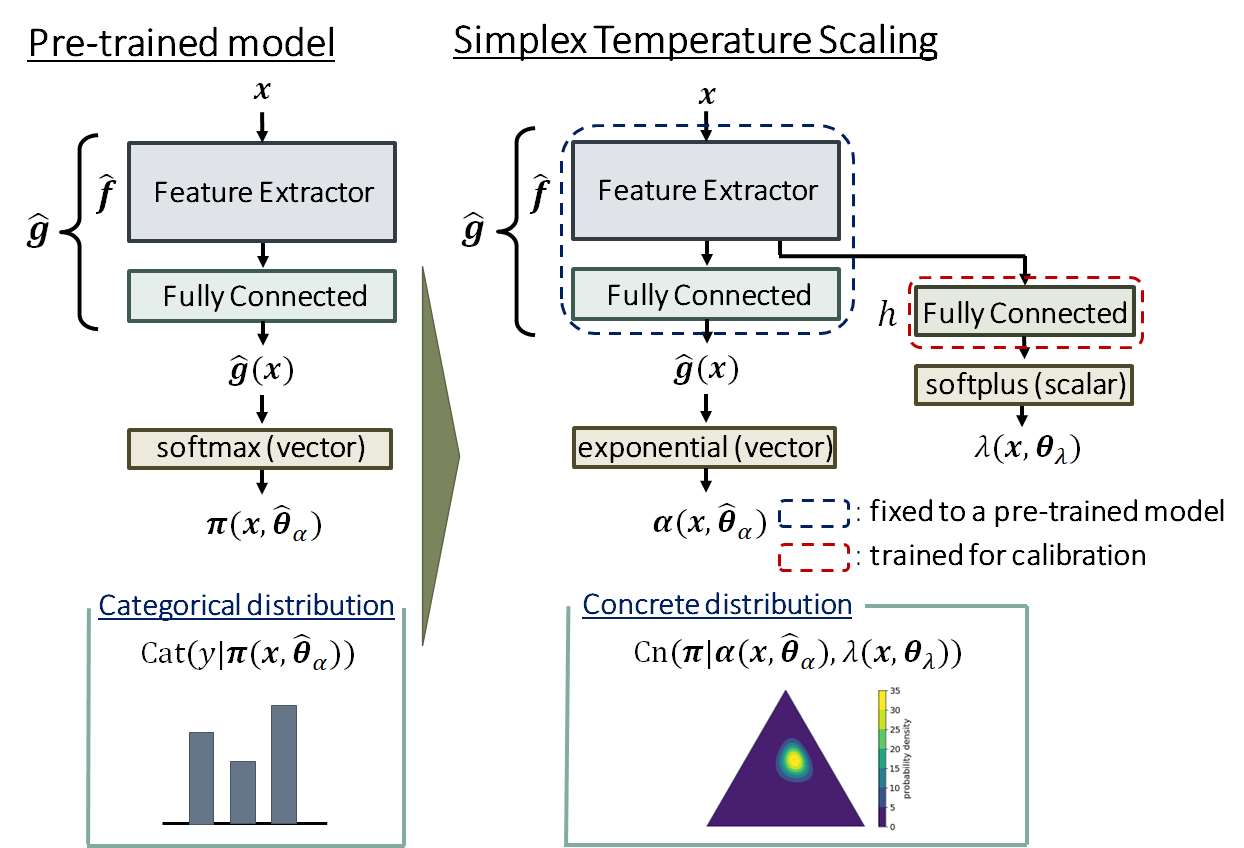}
      \caption{Overview of our proposed method, Simplex Temperature Scaling (STS). The Concrete distribution has two parameters, which are computed using a given pre-trained DNN model and an additional branch.}
      \label{fig:overview}
  \end{figure}
As explained in Section~\ref{sec:location}, we can reuse a pre-trained DNN model for $\bm{\alpha}(\bm{x}, \hat{\bm{\theta}}_{\mathrm{\alpha}})$ and skip Step~1. We compute the location parameter such as
\begin{align}
\bm{\alpha}(\bm{x}, \hat{\bm{\theta}}_{\mathrm{\alpha}})=(e^{\hat{g}_1(\bm{x})},\ldots,e^{\hat{g}_K(\bm{x})})^\top,
\end{align}
where $\hat{\bm{g}} : \mathcal{X}\to\mathbb{R}^K$ is a pre-trained DNN model.
In addition, we introduce another DNN model for $\lambda(\bm{x}, \bm{\theta}_{\mathrm{\lambda}})$ and train it in Step~2. Since many DNN models for classification include a feature extractor~\citep{transfer-learning}, we assume that the DNN model for $\lambda(\bm{x}, \bm{\theta}_{\mathrm{\lambda}})$ shares its feature extractor with $\hat{\bm{g}}$. This sharing helps to reduce the computational complexity of calibration. Let~$\hat{\bm{f}} : \mathcal{X}\to\mathcal{Z}$ be the feature extractor of $\hat{\bm{g}}$, where~$\mathcal{Z}$ is a feature space. In other words, $\hat{\bm{f}}(\bm{x})$ denotes the outputs of a hidden layer in $\hat{\bm{g}}$. The temperature parameter is computed as 
\begin{align}
\lambda(\bm{x}, \bm{\theta}_{\mathrm{\lambda}})=\mathrm{softplus}((h\circ \hat{\bm{f}})(\bm{x})),
\end{align}
where $\mathrm{softplus}(\cdot) : \mathbb{R}\to(0, \infty)$ is the softplus function, which is defined as $\mathrm{softplus}(x)=\ln(1+e^x)$~\citep{softplus}, and $h : \mathcal{Z}\to\mathbb{R}$ is a scalar-valued function computed by fully connected layers.

Although $\bm{\theta}_{\mathrm{\lambda}}$ contains the weight parameters in~$\hat{\bm{f}}$ and~$h$, we optimize only the weight parameters of~$h$, where~$\hat{\bm{f}}$ is fixed, in Step~2.
Figure~\ref{fig:overview} shows the pre-trained DNN model (left side) and the accuracy-preserving calibration by STS with the additional branch $h$ (right side). Based on Corollary~\ref{prop1}, we replace the softmax function of the pre-trained DNN model with the exponential function to obtain $\bm{\alpha}(\bm{x}, \hat{\bm{\theta}}_{\mathrm{\alpha}})$ and add a branch to a hidden layer of the pre-trained DNN model for $\lambda(\bm{x}, \bm{\theta}_{\mathrm{\lambda}})$.
\section{EXPERIMENTS}
\label{sec:ex}
\begin{table*}[t]
    \caption{Expected Calibration Errors (ECEs) and accuracies of the test dataset after calibration by Parameterized Temperature Scaling (PTS)~[ECCV2022]~\citep{PTS}, Adaptive Temperature Scaling (Ada-TS)~[AAAI2023]~\citep{AdaptiveTemp}, and Simplex Temperature Scaling (STS).}
    \label{tab:results}
    \centering
      \begin{tabular}{ccccccc}                                            \toprule
           &      & \multicolumn{4}{c}{ECE (\%)}  &  \multirow{2}{*}{accuracy (\%)}                                                                  \\\cmidrule{3-6}
                                  &                                                    & pre-trained          & PTS & AdaTS & STS (ours)        &        \\\midrule
                       \multirow{2}{*}{FMNIST}   & LeNet5                      &         3.52$\pm$0.15         &  \textbf{0.95$\pm$0.32}     &    \textbf{0.93$\pm$0.28}             &  \textbf{0.91$\pm$0.15}   & 91.2$\pm$0.17        \\
               & ResNet18                      & 3.86$\pm$0.31                   & 2.32$\pm$0.37        & \textbf{1.15}$\pm$\textbf{0.16}                  & \textbf{1.15}$\pm$\textbf{0.28}       & 94.0$\pm$0.15            \\\midrule
        \multirow{2}{*}{CIFAR10}  & VGG16-BN     & 4.40$\pm$0.14                   & 2.48$\pm$0.24        & 1.90$\pm$0.54                   & \textbf{1.48}$\pm$\textbf{0.35}    & 93.7$\pm$0.19         \\
                                  & ResNet18     & 2.80$\pm$0.17                   & \textbf{1.28$\pm$0.18}        & \textbf{1.21$\pm$0.38}                   & \textbf{1.10}$\pm$\textbf{0.24}     & 95.1$\pm$0.14                  \\\midrule
        \multirow{2}{*}{CIFAR100} & ResNet50     & 9.04$\pm$0.36                   & 6.71$\pm$0.56        & \textbf{4.03$\pm$2.04}                  & \textbf{3.73}$\pm$\textbf{1.43}    & 78.7$\pm$0.72          \\
                                  & DenseNet121  & 7.23$\pm$0.59                   & 5.71$\pm$0.81        & 4.22$\pm$1.08                   & \textbf{2.87}$\pm$\textbf{0.25}   & 79.6$\pm$0.50           \\\midrule
        \multirow{2}{*}{STL10}    & VGG16-BN     & 15.23$\pm$0.56                   & 3.84$\pm$0.60      & 4.41$\pm$1.75                   & \textbf{2.15}$\pm$\textbf{0.58}    & 79.4$\pm$0.38                \\
                                  & ResNet18     & 12.88$\pm$0.55                   & 2.01$\pm$0.55        & 2.20$\pm$0.58                   & \textbf{1.48}$\pm$\textbf{0.29}   & 77.8$\pm$0.83          \\\bottomrule
      \end{tabular}
  \end{table*}
The superiority of STS was demonstrated in numerical comparison experiments of STS and existing accuracy-preserving calibration methods.
For the numerical experiments, we selected two recently proposed TS methods: Parameterized Temperature Scaling (PTS)~\citep{PTS} and Adaptive Temperature Scaling (AdaTS)~\citep{AdaptiveTemp}. The same pre-trained DNN model was used in each method and all methods were calibrated. Their calibration performances were then compared. The classification accuracy is shared among all methods. All calculations were performed on a NVIDIA A100 GPU.
\subsection{Datasets}
\label{sec:data}
Experiments were performed on four open datasets for image classification, namely, FashionMNIST (FMNIST)~\citep{FMNIST}, CIFAR10, CIFAR100~\citep{CIFAR}, and STL10~\citep{STL10}. The same datasets were employed in a previous calibration study~\citep{mixup-calibration}.
Each dataset was randomly split into a training dataset, a validation dataset, and a test dataset. The numbers of samples in the training, validation, and test datasets are given in Appendix~\ref{sec:split}.
The models were pre-trained on the training dataset prior to calibration, and the parameters for calibration were tuned on the validation dataset.
While PTS and AdaTS were calibrated using the original samples with hard labels, STS was calibrated by applying Multi-Mixup to the original samples. The calibration performance was evaluated on the test dataset.
\subsection{Setup for Training}
\label{sec:hyper}
We adopted LeNet5~\citep{LeNet}, ResNet18, ResNet50~\citep{ResNet}, VGG16-BN~\citep{VGG}, and DenseNet121~\citep{DenseNet} architectures for the DNN models during pre-training. VGG16-BN defines the VGG16~\citep{VGG} architecture with batch normalization. The pairings between the datasets and architectures are shown in Table~\ref{tab:results}. We applied different architectures for each dataset, depending on the image size and the difficulty of the tasks. During calibration, STS used the pre-trained DNN models for the location parameter, and PTS and AdaTS used them for logits. The architecture of the additional branch for the temperature parameter in STS is explained in Appendix~\ref{sec:model}. The architecture for the temperature parameter in PTS was implemented as described in \citet{PTS}, and that in AdaTS was implemented as in the open source code~\citep{AdaTS-code}.

All architectures on all datasets used the stochastic gradient descent~\citep{SGD} optimizer for pre-training. The learning rate was reduced from 0.1 to 0 by cosine annealing~\citep{CosineAnnealing}. The momentum was 0.9, the weight decay was 0.0005, the batch size was 128, and the number of epochs was 200. The branch for the temperature parameter in STS was trained using the Adam optimizer~\citep{Adam} with a fixed learning rate (0.001), a weight decay of 0.0005, and a batch size of 100. The number of epochs was 500. In each optimization step, 100 samples were generated by Multi-Mixup. The reason for applying the Adam optimizer is to reduce the impact of the learning rate. The temperature parameter in PTS was optimized as described in~\citet{PTS}.
For AdaTS, we used the hyperparameter values set in the open source code~\citep{AdaptiveTemp} to optimize the temperature parameter. 
\subsection{Evaluation}
\label{sec:eval}
To evaluate the calibration performances of the methods, we adopted the Expected Calibration Error (ECE)~\citep{calibration}, which divides samples into multiple bins of confidence levels and measures the difference between the confidence range in each bin and the accuracy of the samples in each bin. A lower ECE indicates a higher calibration performance. 

In the experiments, we computed the accuracy and ECE using the test dataset. The number of bins for ECE was set to 10. In each method, we ran 5 trials with different random seeds and calculated the mean and standard deviation of the accuracies and ECEs over the 5 trials. The number of trials for the pre-training and the subsequent accuracy-preserving calibration are the same, and their random seeds are the same. The partitioning of the training, validation, and test datasets was fixed across the 5 trials.
\subsection{Results}
\label{sec:results}
Table~\ref{tab:results} shows the mean and standard deviation of the ECEs and accuracies in the methods.\footnote{Note that these values may differ from the results in~\citet{PTS} and \citet{AdaptiveTemp}, because the setup of pre-training is different from these previous studies.}
Bold indicates the best results or results within one standard deviation of the best results.
As shown in Table~\ref{tab:results}, STS consistently outperformed the other models, indicating that the performance of STS matches or surpasses the performances of PTS and AdaTS. This result, obtained by estimating the prediction uncertainty from the probabilistic model on the probability simplex, confirms the superiority of STS over PTS and AdaTS for calibration.
\subsection{Accuracy-Preserving Calibration via Dirichlet Distribution}
\label{sec:dirnew} 
As we mentioned in Section~\ref{sec:intro}, one possible approach to realize an accuracy-preserving calibration with a probabilistic model on the probability simplex is combining TS and the Dirichlet distribution.
However, this method has problems as explained in Section~\ref{sec:location}.
We experimentally confirmed that this method caused underconfidence after calibration. In the experiments using CIFAR10 and ResNet18, the mean of ECEs was 67.49 and the standard deviation was 3.84. In Appendix~\ref{sec:dirichlet}, we explain the details of the accuracy-preserving calibration method with the Dirichlet distribution and the experiments to simulate this method.
\subsection{Performance of Out-of-Distribution Detection}
As explained in Section~\ref{sec:uncertainty}, the probabilistic model on the probability simplex can distinguish between aleatoric uncertainty and epistemic uncertainty. To confirm the benefit of this distinction, we performed the experiments of out-of-distribution (OOD) detection~\citep{ood-survay}. The details of the experiments are explained in Appendix~\ref{sec:ood}. In OOD detection, estimating the epistemic uncertainty is important because OOD samples have high epistemic uncertainty for models trained with in-distribution samples. STS achieved higher OOD detection performance than PTS and AdaTS by computing the differential entropy~\citep{Cover} of the estimated Concrete distribution, which is effective in estimating the epistemic uncertainty. The details of the differential entropy are discussed in Appendix~\ref{sec:distinction}.
\section{CONCLUSION}
This paper proposed an accuracy-preserving calibration method called STS, which computes the confidence while considering the prediction uncertainty. STS estimates the prediction uncertainty by applying the Concrete distribution. The pre-trained DNN model requires no update in uncertainty estimation because it optimizes the location parameter of the Concrete distribution regardless of the temperature parameter. We also proposed a method called Multi-Mixup, which generates training samples labeled with vectors on the probability simplex. Multi-Mixup avoids the high training cost of previous methods that must train ensemble models. This advantage of Multi-Mixup extends to uncertainty estimation methods other than STS.

\appendix
\onecolumn

\section{MISSING PROOFS}
\label{sec:proof}
This section describes the proofs missing from the main paper.
\subsection{Proof of Theorem \ref{prop2}}
\label{sec:proof2}
By Eqs.~\eqref{eq:ind} and~\eqref{eq:con}, the following holds, where $\mathbb{E}[\cdot]$ denotes the expectation of a stochastic variable $\bm{\pi}\sim \mathrm{Cn}(\bm{\pi}|\bm{\alpha}(\bm{x}, \bm{\theta}_{\mathrm{\alpha}}), \lambda(\bm{x}, \bm{\theta}_{\mathrm{\lambda}}))$ and $\mathbb{P}(\cdot)$ returns the probability that the event in the parentheses occurs under the stochastic variable $\bm{\pi}\sim \mathrm{Cn}(\bm{\pi}|\bm{\alpha}(\bm{x}, \bm{\theta}_{\mathrm{\alpha}}), \lambda(\bm{x}, \bm{\theta}_{\mathrm{\lambda}}))$.
\begin{align}
    p(y=k|\bm{x})&=\int p(y=k|\bm{\pi})p(\bm{\pi}|\bm{x})d\bm{\pi}\\
        &=\int \mathds{1}_{\{\mathrm{arg}\max_{i\in\mathcal{Y}}\pi_i=k\}}\mathrm{Cn}(\bm{\pi}|\bm{\alpha}(\bm{x}, \bm{\theta}_{\mathrm{\alpha}}),  \lambda(\bm{x}, \bm{\theta}_{\mathrm{\lambda}}))d\bm{\pi} \label{eq:cc}\\
        &=\mathbb{E}\left[\mathds{1}_{\{\mathrm{arg}\max_{i\in\mathcal{Y}}\pi_i=k\}}\right]\\
        &=\mathbb{P}\left(\argmax_{i\in\mathcal{Y}}\pi_i=k\right) \label{eq:expect}\\
        &=\frac{\alpha_k(\bm{x}, \bm{\theta}_{\mathrm{\alpha}})}{\sum^K_{i=1}\alpha_i(\bm{x}, \bm{\theta}_{\mathrm{\alpha}})}\ \ \ \mbox{for}\ \forall{k}\in\mathcal{Y}. \label{eq:previous}
\end{align}
Equation~\eqref{eq:cc} holds by Eqs.~\eqref{eq:ind} and~\eqref{eq:con}.
Equation~\eqref{eq:expect} holds by the definition of the expectation.
Equation~\eqref{eq:previous} holds under Proposition 1(b) in~\citet{concrete}. \qed
\subsection{Proof of Corollary \ref{prop1}}
\label{sec:proof1}
Let $\bm{\pi}(\bm{x}, \bm{\theta}_{\mathrm{\alpha}})$ be defined as follows.
\begin{align}
\bm{\pi}(\bm{x}, \bm{\theta}_{\mathrm{\alpha}})\coloneqq \left(\frac{\alpha_1(\bm{x}, \bm{\theta}_{\mathrm{\alpha}})}{\sum^K_{k=1}\alpha_k(\bm{x}, \bm{\theta}_{\mathrm{\alpha}})},\ldots,\frac{\alpha_K(\bm{x}, \bm{\theta}_{\mathrm{\alpha}})}{\sum^K_{k=1}\alpha_k(\bm{x}, \bm{\theta}_{\mathrm{\alpha}})}\right)^\top. \label{eq:wowow}
\end{align}
From Theorem~\ref{prop2}, Eq.~\eqref{eq:predictive} is described as follows, under Eqs.~\eqref{eq:ind} and~\eqref{eq:con}.
\begin{align}
    p(y=k|\bm{x})=\frac{\alpha_k(\bm{x}, \bm{\theta}_{\mathrm{\alpha}})}{\sum^K_{i=1}\alpha_i(\bm{x}, \bm{\theta}_{\mathrm{\alpha}})} 
        =\pi_k(\bm{x}, \bm{\theta}_{\mathrm{\alpha}})\ \ \ \mbox{for}\ \forall{k}\in\mathcal{Y}. \label{eqp}
\end{align}
Therefore, optimizing~$\bm{\theta}_{\mathrm{\alpha}}$ on $\mathcal{D}$ using the maximum likelihood estimation of Eq.~\eqref{eq:predictive} is equivalent to optimizing~$\bm{\theta}_{\mathrm{\alpha}}$ on $\mathcal{D}$ using the maximum likelihood estimation of $\mathrm{Cat}(y|\bm{\pi}(\bm{x}, \bm{\theta}_{\mathrm{\alpha}}))$ $\cdots$(A).

Meanwhile, since $\bm{\alpha}(\bm{x}, \bm{\theta}_{\mathrm{\alpha}})=(e^{g_1(\bm{x})},\ldots,e^{g_K(\bm{x})})^\top$, Eq.~\eqref{eq:cross} is also described as follows.
\begin{align}
\mathcal{L}(\bm{\theta}_{\mathrm{\alpha}}) = - \frac{1}{N}\sum^N_{n=1}\ln \mathrm{Cat}(y_n|\bm{\pi}(\bm{x}_n, \bm{\theta}_{\mathrm{\alpha}})). \label{eqcat}
\end{align}
Equation~\eqref{eqcat} is the negative log-likelihood of $\mathrm{Cat}(y|\bm{\pi}(\bm{x}, \bm{\theta}_{\mathrm{\alpha}}))$. Therefore, optimizing~$\bm{\theta}_{\mathrm{\alpha}}$ on $\mathcal{D}$ using the minimization of Eq.~\eqref{eq:cross} is equivalent to optimizing~$\bm{\theta}_{\mathrm{\alpha}}$ on $\mathcal{D}$ using the maximum likelihood estimation of $\mathrm{Cat}(y|\bm{\pi}(\bm{x}, \bm{\theta}_{\mathrm{\alpha}}))$ $\cdots$(B).

From (A) and (B), we can prove Corollary~\ref{prop1}. \qed
\newpage
\section{ALGORITHM FOR MULTI-MIXUP}
\label{sec:alg}

Algorithm~\ref{alg1} shows the details of Multi-Mixup in each optimization step. We transform samples with one-hot labels in classification datasets to samples with labels on the probability simplex that indicate the prediction uncertainty.
At each optimization step, we randomly sample $K$ mini-batches $\mathcal{B}^{(1)},\ldots,\mathcal{B}^{(K)}$, each containing samples in each class, from the classification dataset. Random sampling is performed so that all mini-batches have the same number of samples, which is denoted by $R\in\mathbb{N}$.\footnote{Even if we use a class-imbalanced dataset, we still sample equally across classes. In this case, we upsample for classes with small sample sizes. The evaluation of STS in a class-imbalanced situation is left as future work.}
In Multi-Mixup, we linearly interpolate the $R\times K$ samples among classes~(Lines~6-9), which is repeated $S$ times. For each $s\in\{1,\ldots,S\}$ and $r\in\{1,\ldots,R\}$, $\bm{y}^{(k)}_{s,r}$ denotes a one-hot vector whose $k$-th entry is 1 and whose other entries are 0. In each of the $S$ iterations, we generate the value of the weight~$\bm{w}$ by randomly sampling from the Dirichlet distribution~$\mathrm{Dir}(\bm{\pi}|\beta\bm{1}_{K})$ (Line~2), where~$\bm{1}_{K}$ is a $K$-dimensional vector whose all entries are~1 and $\beta$ is a hyperparameter in $(0, \infty)$. In addition, we randomly shuffle the order of samples in each mini-batch (Lines~3-5). In the experiments explained in Section~\ref{sec:ex}, we generated 100 samples at each step of the optimization by setting $R=10$ and $S=10$ in Alg.~\ref{alg1}. Other settings are discussed in Appendix~\ref{sec:RS}.
{
\setlength{\algomargin}{1.3em}
\begin{algorithm}[t]
    \caption{Algorithm to synthetically generate a dataset in which the inputs are labeled with vectors on the probability simplex by Multi-Mixup.}
    \label{alg1}
    \KwInput{$\mathcal{B}^{(k)}(k\in\mathcal{Y})$: mini-batch including the pairs of inputs and one-hot labels in each class ($|\mathcal{B}^{(1)}|=\cdots=|\mathcal{B}^{(K)}|=R$, $R\in\mathbb{N}$)\hspace{10cm} $\beta\in(0, \infty)$: hyperparameter to generate the weight $\bm{w}$ by random sampling\hspace{6cm} $S\in\mathbb{N}$: the number of iterations to randomly shuffle the samples.}
    \KwOutput{$\tilde{\mathcal{B}}$: mini-batch generated by Multi-Mixup}
    
    \For{$s=1,\ldots,S$}{
        $\bm{w}\sim \mathrm{Dir}(\bm{\pi}|\beta\bm{1}_{K})$
        
        \For{$k=1,\ldots,K$}{
            Randomly shuffle the samples in $\mathcal{B}^{(k)}$, and get $\mathcal{B}^{(k)}_s=\{(\bm{x}^{(k)}_{s,r}, \bm{y}^{(k)}_{s,r})\}^{R}_{r=1}$
        }
        \For{$r=1,\ldots,R$}{
            
            $\tilde{\bm{x}}_{s,r}\leftarrow\sum^K_{k=1}w_k\bm{x}^{(k)}_{s,r}$

            $\tilde{\bm{\pi}}_{s,r}\leftarrow\sum^K_{k=1}w_k\bm{y}^{(k)}_{s,r}$
        }
    }
    \Return $\tilde{\mathcal{B}}=\{(\tilde{\bm{x}}_{s,r}, \tilde{\bm{\pi}}_{s,r})\}_{s=1,\ldots,S,r=1,\ldots,R}$
\end{algorithm}
}

Note that, in Line~8, $\bm{y}^{(i)}_{s,r}\neq \bm{y}^{(j)}_{s,r}$ for $i\neq j$.
If we use a different policy than Alg.~\ref{alg1} and randomly select~$K$ samples without distinguishing the classes, samples corresponding to the surfaces and edges of~$\Delta_{K-1}$ would be generated when there are class overlaps among the samples. In this case, if the number of classes~$K$ is large, the proportion of samples on the interior of~$\Delta_{K-1}$ becomes extremely small, and there is a risk of obtaining a model that only returns a temperature parameter value close to 0 after training. Therefore, we formulate Multi-Mixup so that samples in different classes are interpolated.

The random sampling of~$\bm{w}$ by $\mathrm{Dir}(\bm{\pi}|\beta\bm{1}_{K})$ is based on the fact that Mixup~\citep{mixup} determines the weight of the interpolation by sampling from the beta distribution~$\mathrm{Beta}(\pi|\beta, \beta)$.
Although we can also generate $\bm{w}$ by sampling from the Concrete distribution, a temperature parameter value that represents a uniform distribution on~$\Delta_{K-1}$ varies with the number of dimensions, $K$, and we cannot describe this value in a closed form~\citep{concrete}. 
Conversely, the Dirichlet distribution can represent a uniform distribution on~$\Delta_{K-1}$ regardless of~$K$ if we specify $\beta=1$.
In the experiments explained in Section~\ref{sec:ex}, we varied the value of~$\beta$ from 0.2 to 2.0 in 0.1 increments and tried STS for each setting. Then, we calculated ECE for each setting with the validation dataset\footnote{The validation dataset was used for both training the temperature parameter and tuning the hyperparameter.} and selected the setting that gave the smallest ECE. Since there were some settings where the loss shown in Eq.~\eqref{eq:loss} diverged during optimization, such settings were automatically excluded. Finally, we evaluated the selected setting on the test dataset.
The sensitivity of $\beta$ to ECE is discussed in Appendix~\ref{sec:sensitive}.
\section{SUPPLEMENTS OF EXPERIMENTS IN THE MAIN PAPER}
\label{sec:supp}
\begin{table}[t]
    \caption{Number of samples in each dataset.}
    \label{tab:dataset}
    \centering
      \begin{tabular}{cccc}                                            \toprule
                                & training & validation & test \\\midrule
        FashionMNIST~\citep{FMNIST} & 60000      & 5000       & 5000         \\
        CIFAR10~\citep{CIFAR}               & 50000      & 5000       & 5000         \\
        CIFAR100~\citep{CIFAR}              & 50000      & 5000       & 5000         \\
        STL10~\citep{STL10}                 & 5000       & 4000       & 4000         \\\bottomrule
      \end{tabular}
  \end{table}

\begin{figure}[t]
\begin{minipage}{0.49\columnwidth}
    \centering
      \includegraphics[width=6.5cm, height=4.5cm]{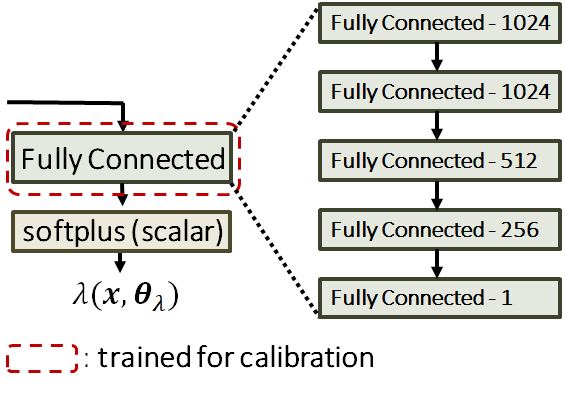}
      \caption{Architecture of the additional branch for the temperature parameter $\lambda(\bm{x}, \bm{\theta}_{\mathrm{\lambda}})$ in the experiments. We used the same architecture regardless of the architecture of the pre-trained DNN model.}
      \label{fig:fully}
\end{minipage}
\begin{minipage}{0.49\columnwidth}
\centering
      \includegraphics[width=6cm, height=4.5cm]{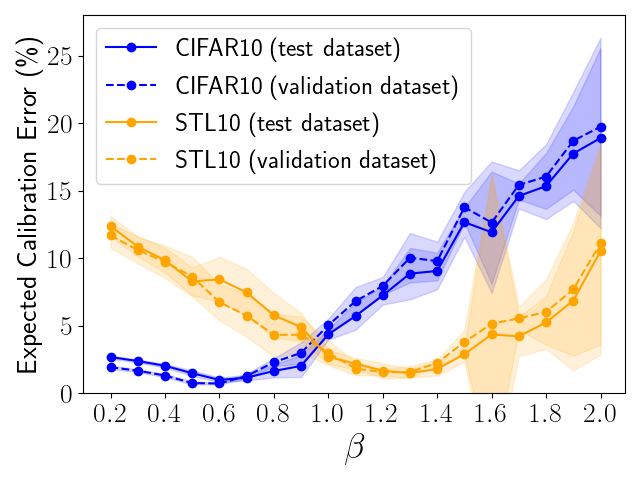}
      \caption{Transitions of Expected Calibration Errors (ECEs) when we varied the value of $\beta$ from 0.2 to 2.0 in 0.1 increments. Each point indicates the median of 5 trials and the shading indicates the standard deviation.}
      \label{fig:hyper}
\end{minipage}
\end{figure}

\subsection{Dataset}
\label{sec:split}
Table~\ref{tab:dataset} shows the number of samples in the training, validation, and test datasets used in the experiments explained in Section~\ref{sec:ex}. The original files of FasionMNIST, CIFAR10, and CIFAR100 contain a training dataset and a test dataset, respectively. For the validation dataset, we randomly split the original test dataset into two datasets. The original file of STL10 contains a training dataset, a test dataset, and an unlabeled dataset. In this study, we did not use the unlabeled dataset and used only the training and test datasets. Similar to other datasets, we randomly split the original test dataset into two datasets for the validation dataset. The characteristic of STL10 is a small number of training samples, which causes overfitting to the training dataset. This characteristic is reflected in the test accuracy shown in Table~\ref{tab:results}.

\subsection{Setup of the Models}
\label{sec:model}
As explained in Section~\ref{sec:archit}, to compute the temperature parameter, we used an additional branch different from a pre-trained DNN model. Figure~\ref{fig:fully} shows the architecture of the branch to compute the temperature parameter. The same architecture was applied to all datasets and architectures of the pre-training. For the STL10 and VGG16-BN pairing only, we reduced the feature dimension from 512$\times$3$\times$3 to 512 by average pooling~\citep{pooling}. This process is due to the high resolution of STL10. The experiments with a different architecture are discussed in Appendix~\ref{sec:arc}.

\subsection{Sensitivity of the Hyperparameter in Multi-Mixup}
\label{sec:sensitive}
As explained in Appendix~\ref{sec:alg}, we selected the best value of the hyperparameter in Multi-Mixup, $\beta$, by the validation dataset. In this regard, we confirmed the difference in calibration performance among the tried values of~$\beta$. Figure~\ref{fig:hyper} shows the ECEs when we set 0.2-2.0 as the value of $\beta$. We chose the experiments with CIFAR10 and STL10 and used the models calibrated by STS with different values of $\beta$. The pre-training architecture was ResNet18. As can be seen in Fig.~\ref{fig:hyper}, the different datasets show different trends. While the best value of $\beta$ is 0.6 on CIFAR10, that is 1.3 on STL10. However, the ECEs calculated on the validation dataset are consistent with those on the test dataset. This means that tuning $\beta$ on the validation dataset results in high calibration performance on the test dataset, although this tuning is required on a per-task basis.
\subsection{Time Required for Training and Inference}
As part of a comparison with existing methods, we measured the time to calibrate confidence, which corresponds to training, and the time to compute confidence after calibration, which corresponds to inference, for PTS, AdaTS, and STS. We chose the experiments with the pair of CIFAR10 and ResNet18. The number of training epochs was set as described in Section~\ref{sec:hyper}. The time measurements were performed on a Xeon Platinum 8358 CPU. The number of workers in DataLoader~\citep{PyTorch} and that of cores were unified to 8. We used a NVIDIA A100 GPU as described in the main paper. Table~\ref{tab:time} shows the results of the time measurements. The time required for training varies greatly from method to method. This variation is caused by the difference in the objective function used to optimize the temperature parameter and the difference in the number of epochs due to the objective function. Although STS is shorter than PTS, that is longer than AdaTS. Therefore, we should devise a method to shorten the training time in future work. In contrast, there is little difference in the time required for inference among the methods. This implies that the sampling from the Gumbel distribution required to compute our confidence in STS does not hurt the running time compared to the existing methods.
\begin{table}[t]
    \caption{Time to calibrate confidence (training) and the time to compute confidence (inference).}
    \label{tab:time}
    \centering
      \begin{tabular}{cccccc}                                            \toprule
           & PTS & AdaTS  & STS (ours)   \\\midrule
         training (sec.)  &  26142.47 &  269.88 & 5097.18 \\
         inference  (sec.)& 18.37 & 18.03  & 17.91  \\\bottomrule
      \end{tabular}
  \end{table}
\section{INAPPROPRIATENESS OF DIRICHLET DISTRIBUTION FOR ACCURACY-PRESERVING CALIBRATION}
\label{sec:dirichlet}
\subsection{Calibration with Dirichlet Distribution}
\label{sec:dirm}
In this section, we explain why the probabilistic model as in Eq.~\eqref{eq:dir} is inappropriate for accuracy-preserving calibration in order to reinforce the validity of adopting the Concrete distribution.
We consider calibration using Eq.~\eqref{eq:dir} as in previous studies~\citep{PriorNet,EnD2,generalEnD2,ScalingEnD2}. Moreover, we add a temperature parameter into the parameter of the Dirichlet distribution as 
\begin{align}
\bm{\mu}(\bm{x}, \bm{\theta}_{\mathrm{\mu}})=\left(\exp\left(\frac{\hat{g}_1(\bm{x})} {t(\bm{x},\bm{\theta}_{\mathrm{\mu}})}\right), \cdots, \exp\left(\frac{\hat{g}_K(\bm{x})} {t(\bm{x},\bm{\theta}_{\mathrm{\mu}})}\right)\right)^\top, \label{eql}
\end{align}
where $\hat{\bm{g}} : \mathcal{X}\to\mathbb{R}^K$ is a pre-trained DNN model and $t(\cdot, \bm{\theta}_{\mathrm{\mu}}) : \mathcal{X}\to(0, \infty)$ is a scalar-valued function parameterized by a real vector $\bm{\theta}_{\mathrm{\mu}}$. We determined the position of~$t(\bm{x},\bm{\theta}_{\mathrm{\mu}})$ in Eq.~\eqref{eql} based on the temperature annealing discussed in~\citet{EnD2}.
In this case, the negative log-likelihood of $\mathrm{Dir}(\bm{\pi}|\bm{\mu}(\bm{x}, \bm{\theta}_{\mathrm{\mu}}))$ is formulated as
\begin{align}
    \begin{split}
      \mathcal{L}(\bm{\theta}_{\mathrm{\mu}})
      & = - \frac{1}{M}\sum^M_{m=1}\ln \Gamma\left(\sum^K_{k=1} \exp\left(\frac{\hat{g}_k(\tilde{\bm{x}}_m)} {t(\tilde{\bm{x}}_m,\bm{\theta}_{\mathrm{\mu}})}\right)\right)+\frac{1}{M}\sum^M_{m=1}\sum^K_{k=1}\ln \Gamma\left( \exp \left( \frac{\hat{g}_k(\tilde{\bm{x}}_m)}{t(\tilde{\bm{x}}_m,\bm{\theta}_{\mathrm{\mu}})}\right)\right) \\
      &\hspace{5cm} - \frac{1}{M}\sum^M_{m=1}\sum^K_{k=1}\left(\exp\left(\frac{\hat{g}_k(\tilde{\bm{x}}_m)}{t(\tilde{\bm{x}}_m, \bm{\theta}_{\mathrm{\mu}})}\right)-1\right)\ln \tilde{\pi}_{m,k}, \label{eqdiri}
    \end{split}
  \end{align}
where $\Gamma(\cdot)$ is the gamma function and $\{(\tilde{\bm{x}}_m, \tilde{\bm{\pi}}_m)\}^M_{m=1}$ denotes the dataset generated by Multi-Mixup.
For an accuracy-preserving calibration, the temperature parameter is computed as $t(\bm{x}, \bm{\theta}_{\mathrm{\mu}})=\mathrm{softplus}((h\circ \hat{\bm{f}})(\bm{x}))$, where $h : \mathcal{Z}\to\mathbb{R}$ is a scalar-valued function computed by fully connected layers. The weight parameters of $h$ are optimized by minimizing Eq.~\eqref{eqdiri}, where $\hat{\bm{g}}$ and $\hat{\bm{f}}$ are fixed.

Let $t(\bm{x}, \hat{\bm{\theta}}_{\mathrm{\mu}})$ be the temperature parameter estimated by Eq.~\eqref{eqdiri}. We compute the confidence as follows.
\begin{align}
    \mathrm{confidence}(\bm{x})=\max_{k\in \mathcal{Y}}(\mathbb{E}[\bm{\pi}])_k =\max_{k\in\mathcal{Y}}\frac{\exp\left(\frac{\hat{g}_k(\bm{x})}{t(\bm{x}, \hat{\bm{\theta}}_{\mathrm{\mu}})}\right)}{\sum^K_{i=1}\exp\left(\frac{\hat{g}_i(\bm{x})}{t(\bm{x}, \hat{\bm{\theta}}_{\mathrm{\mu}})}\right)},\label{eqconfidence}
\end{align}
where $\mathbb{E}[\cdot]$ denotes the expectation of $\mathrm{Dir}(\bm{\pi}|\bm{\mu}(\bm{x}, \hat{\bm{\theta}}_{\mathrm{\mu}}))$. This previous study takes advantage of the fact that the Dirichlet distribution is a conjugate prior, and Eq.~\eqref{eqconfidence} is equivalent to the maximum of Eq.~\eqref{eq:predictive}.

Using the Dirichlet distribution, the maximum likelihood estimation of Eq.~\eqref{eq:predictive} is equivalent to the training with the cross-entropy loss similar to STS. Therefore, if we denote $\bm{\mu}_{\hat{\bm{g}}}(\bm{x}, \bm{\theta}_{\mathrm{\mu}})=(e^{\hat{g}_1(\bm{x})},\ldots,e^{\hat{g}_K(\bm{x})})^\top$, $\bm{\mu}_{\hat{\bm{g}}}(\bm{x}, \bm{\theta}_{\mathrm{\mu}})$ is optimal in terms of the likelihood of Eq.~\eqref{eq:predictive}. However, the optimality of this parameter is not guaranteed after training $t(\cdot, \bm{\theta}_{\mathrm{\mu}})$ solely using Eq.~\eqref{eqdiri}. In other words, training $t(\cdot,\bm{\theta}_{\mathrm{\mu}})$ by Eq.~\eqref{eqdiri} may make $\bm{\mu}_{\hat{\bm{g}}}(\bm{x}, \bm{\theta}_{\mathrm{\mu}})$ less optimal in terms of the cross-entropy loss. This problem results in poor calibration performance. We confirmed that the accuracy-preserving calibration with Eq.~\eqref{eqdiri} leads to underconfidence by numerical experiments. The experiments are explained in the following section.
  
\subsection{Experiments to Confirm Inappropriateness}
\begin{figure}[t]
    \begin{minipage}{0.49\columnwidth}
      \centering
        \includegraphics[width=4.8cm, height=4cm]{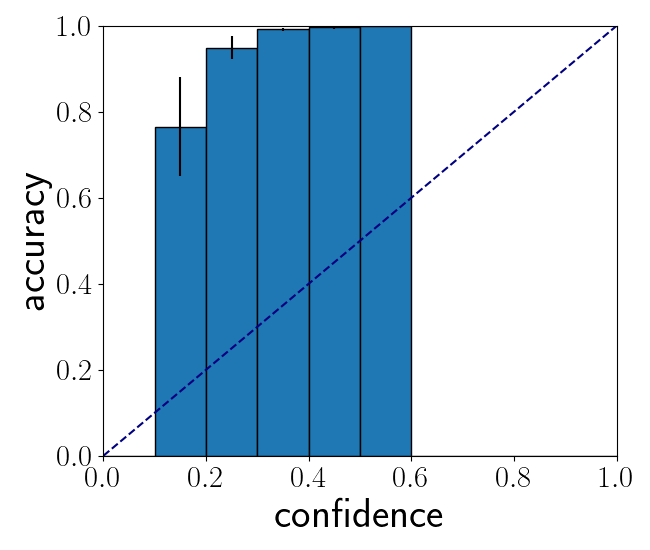}
        \subcaption{Dirichlet distribution.}
        \label{fig:dirichlet}
    \end{minipage}
    \begin{minipage}{0.49\columnwidth}
      \centering
        \includegraphics[width=4.8cm, height=4cm]{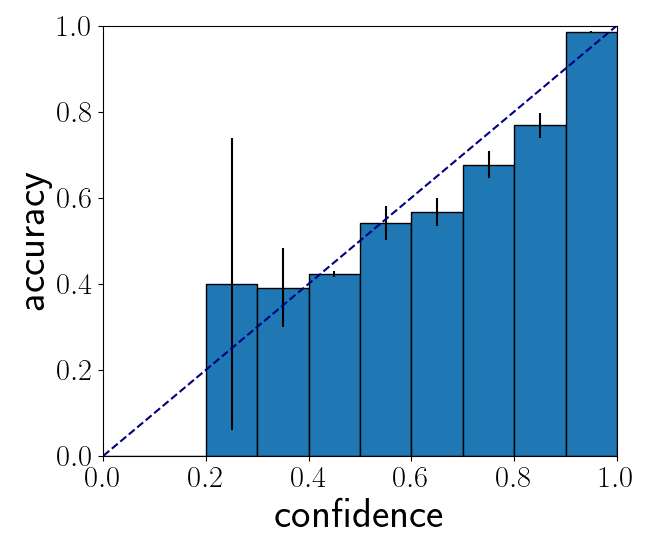}
        \subcaption{Concrete distribution.}
        \label{fig:concrete}
    \end{minipage}
    \caption{Relation between the confidence and classification accuracy when we use the probabilistic model as in Eq.~\eqref{eq:dir} (left side) and the probabilistic model as in Eqs.~\eqref{eq:ind} and ~\eqref{eq:con} (right side). The interval between 0 and 1 is divided into 10 bins, and the height of the bars indicates the classification accuracy of the samples whose confidence is within each bin. The bins without a bar imply that there is no sample in the test dataset that has a confidence level within the corresponding bins. We plotted the mean of 5 trials with different random seeds, and the standard deviation is represented by error bars. The diagonal dashed line represents the ideal case where confidence and classification accuracy are in perfect agreement, and the closer the height of the bars is to the dashed line, the more accurate the calibration.}\label{fig:curve}
  \end{figure}

  \begin{figure}[t]
    \begin{minipage}{0.19\columnwidth}
      \centering
        \includegraphics[width=3.5cm, height=2.3cm]{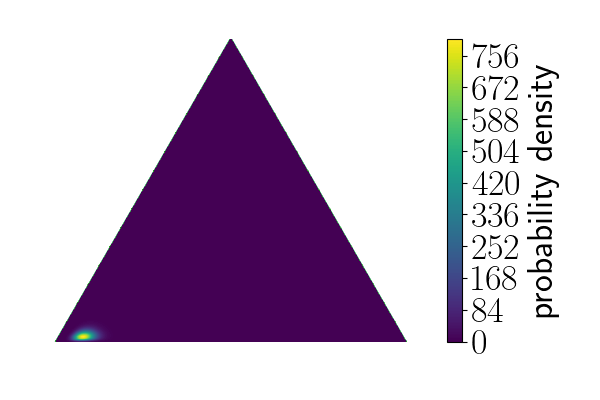}
        \subcaption{$t(\bm{x}, \bm{\theta}_{\mathrm{\mu}})=0.2$}
      
    \end{minipage}
    \begin{minipage}{0.19\columnwidth}
      \centering
        \includegraphics[width=3.3cm, height=2.5cm]{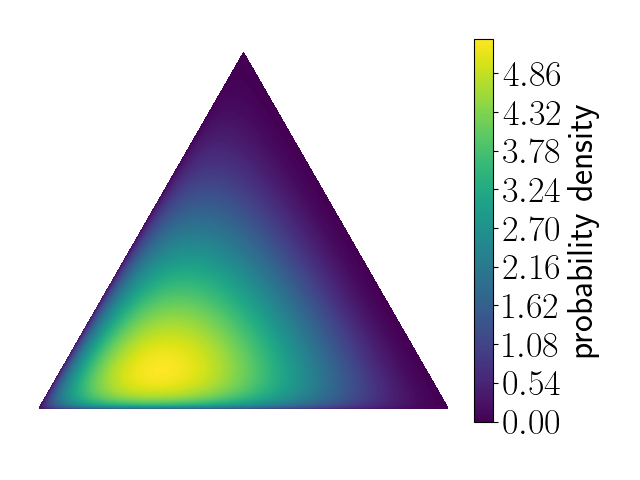}
        \subcaption{$t(\bm{x}, \bm{\theta}_{\mathrm{\mu}})=1$}
      
    \end{minipage}
     \begin{minipage}{0.19\columnwidth}
      \centering
        \includegraphics[width=3.3cm, height=2.5cm]{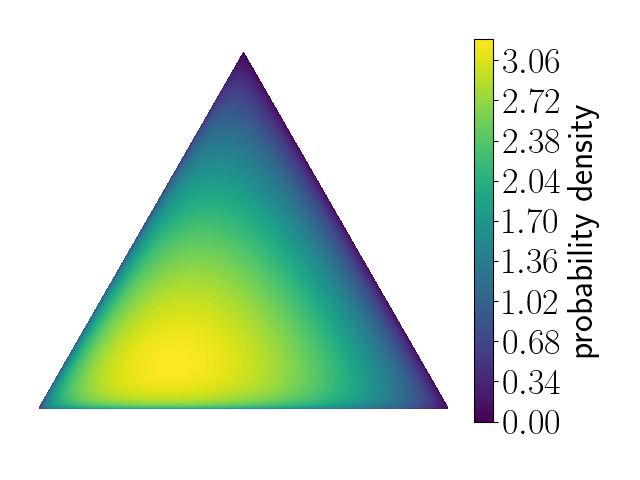}
        \subcaption{$t(\bm{x}, \bm{\theta}_{\mathrm{\mu}})=2$}
      
    \end{minipage}
    \begin{minipage}{0.19\columnwidth}
      \centering
        \includegraphics[width=3.3cm, height=2.5cm]{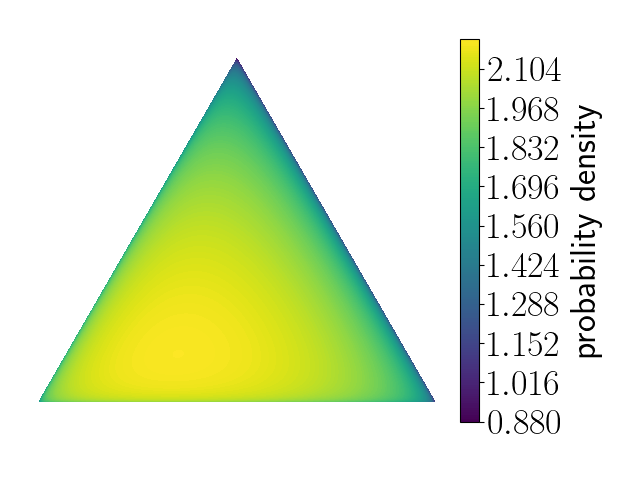}
        \subcaption{$t(\bm{x}, \bm{\theta}_{\mathrm{\mu}})=10$}
      
    \end{minipage}
    \begin{minipage}{0.19\columnwidth}
      \centering
        \includegraphics[width=3.3cm, height=2.5cm]{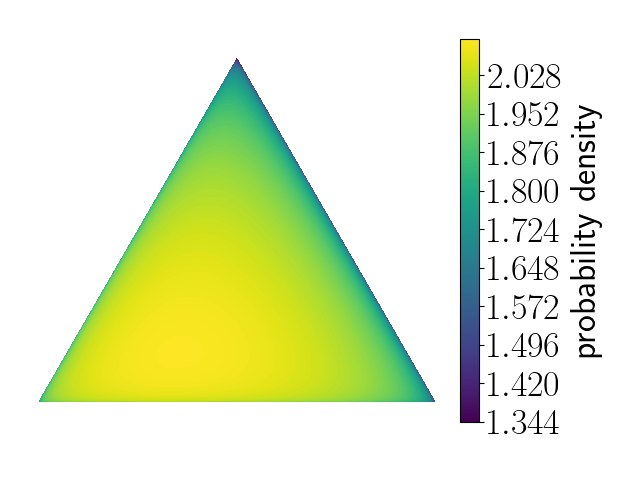}
        \subcaption{$t(\bm{x}, \bm{\theta}_{\mathrm{\mu}})=20$}
      
    \end{minipage}
    \caption{Variation of the Dirichlet distribution when we shifted only $t(\bm{x}, \bm{\theta}_{\mathrm{\mu}})$ with fixed $\hat{\bm{g}}(\bm{x})$. In this figure, we did not use any input $\bm{x}$ and give pseudo values for $t(\bm{x}, \bm{\theta}_{\mathrm{\mu}})$ and $\hat{\bm{g}}(\bm{x})$. The value of $\hat{\bm{g}}(\bm{x})$ was fixed to $(1.0, 0.5, 0.25)^\top$. The value of $t(\bm{x}, \bm{\theta}_{\mathrm{\mu}})$ was shifted from 0.2 to 20.}\label{fig:dir}
    \begin{minipage}{0.19\columnwidth}
      \centering
        \includegraphics[width=3.3cm, height=2.5cm]{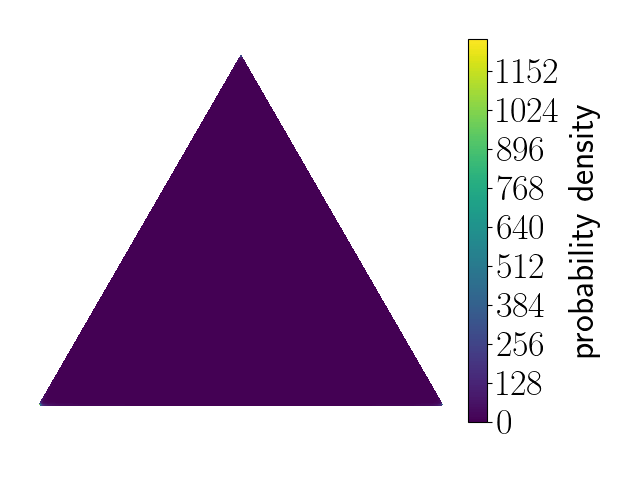}
        \subcaption{$\lambda=0.2$}\label{a}
      
    \end{minipage}
    \begin{minipage}{0.19\columnwidth}
      \centering
        \includegraphics[width=3.3cm, height=2.5cm]{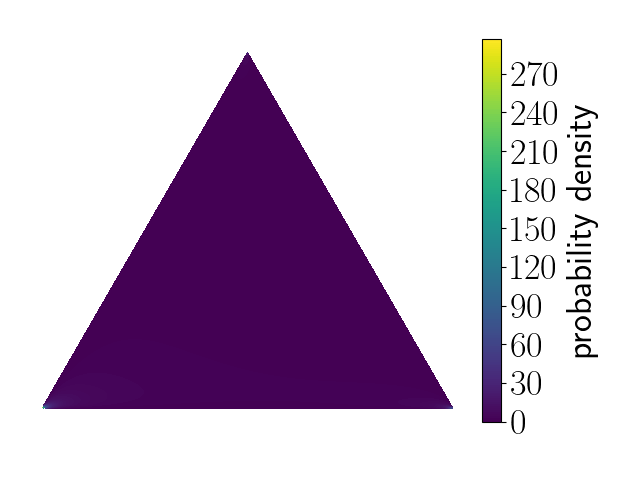}
        \subcaption{$\lambda=1$}\label{b}
      
    \end{minipage}
     \begin{minipage}{0.19\columnwidth}
      \centering
        \includegraphics[width=3.3cm, height=2.5cm]{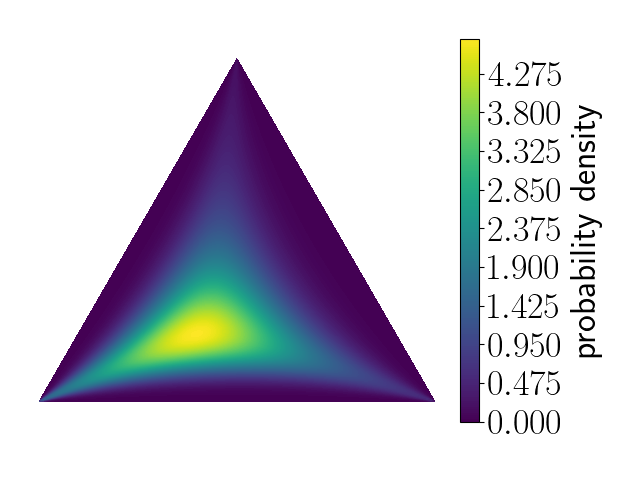}
        \subcaption{$\lambda=2$}
      
    \end{minipage}
    \begin{minipage}{0.19\columnwidth}
      \centering
        \includegraphics[width=3.3cm, height=2.5cm]{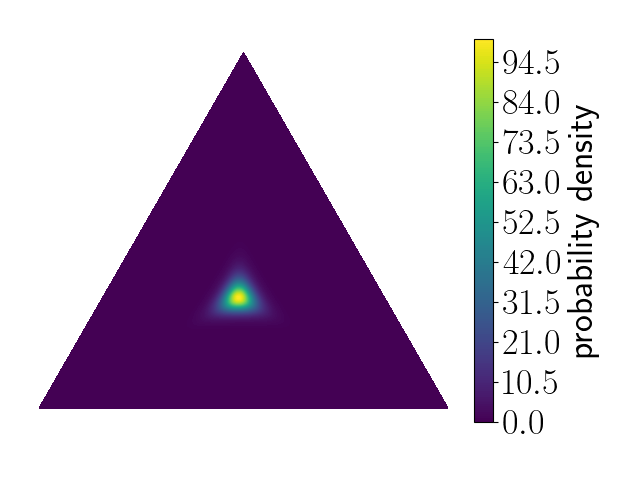}
        \subcaption{$\lambda=10$}
      
    \end{minipage}
    \begin{minipage}{0.19\columnwidth}
      \centering
        \includegraphics[width=3.3cm, height=2.5cm]{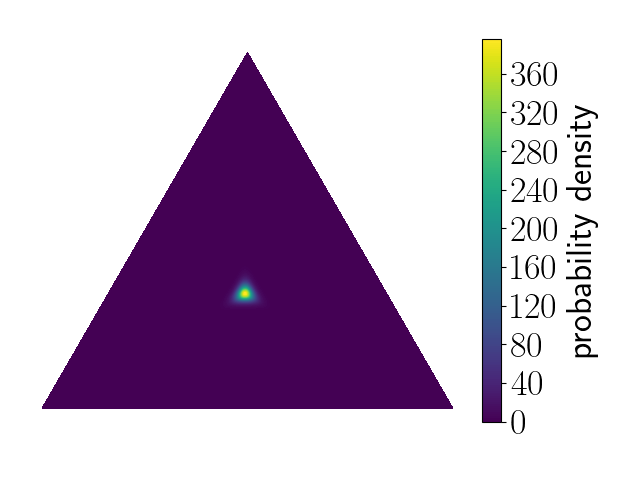}
        \subcaption{$\lambda=20$}
      
    \end{minipage}
    \caption{Variation of the Concrete distribution when we shifted only the temperature parameter $\lambda$ with the fixed location parameter $\bm{\alpha}$. The value of $\bm{\alpha}$ was fixed to $(e^{1.0}, e^{0.5}, e^{0.25})^\top$. The value of $\lambda$ was shifted from 0.2 to 20. Due to the high probability density, bright areas are almost non-existent in Figs.~\ref{a} and~\ref{b}. In these figures, the probability density is concentrated at the lower left apex.}\label{fig:con}
  \end{figure}
We used CIFAR10~\citep{CIFAR} as a dataset and used ResNet18~\citep{ResNet} as a model architecture. The pre-trained DNN model $\hat{\bm{g}}$ was the same as in Section~\ref{sec:ex}. The architecture for the temperature parameter $t(\cdot, \bm{\theta}_{\mathrm{\mu}})$ and the setup for training $t(\cdot, \bm{\theta}_{\mathrm{\mu}})$ and tuning the hyperparameter $\beta$ were the same as in STS, except for the learning rate.
The learning rate was set to $10^{-6}$ because the loss function~\eqref{eqdiri} diverged when set to the same value as in STS. After training $t(\cdot, \bm{\theta}_{\mathrm{\mu}})$, we computed ECE with the test dataset.

As a result, the mean of ECE was 67.49 and the standard deviation was 3.84, which is worse than the pre-trained DNN model.
From this result, we can experimentally confirm that the Dirichlet distribution is inappropriate for accuracy-preserving calibration even if we add a temperature parameter to the Dirichlet distribution. Figure~\ref{fig:dirichlet} implies the cause of the poorer ECE. We have plotted the relation between the confidence computed by Eq.~\eqref{eqconfidence} and the classification accuracy. For reference, we also present the same type of figure for STS, in Fig.~\ref{fig:concrete}. From Fig.~\ref{fig:dirichlet}, the accuracy-preserving calibration using the Dirichlet distribution results in confidence levels of only 0.6 or less although the classification accuracy of the pre-trained DNN model is about 95\% as shown in Table~\ref{tab:results}. In other words, extreme underconfidence is observed.

As we mentioned in the main paper, the failure of the accuracy-preserving calibration when we used the Dirichlet distribution is caused by the fact that the maximality of the likelihood of Eq.~\eqref{eq:predictive} (i.e., the minimality of the cross-entropy loss) does not hold when only $t(\cdot, \bm{\theta}_{\mathrm{\mu}})$ is trained using Eq.~\eqref{eqdiri}.
Moreover, even if only $t(\cdot, \bm{\theta}_{\mathrm{\mu}})$ increases with fixed $\hat{\bm{g}}$, the Dirichlet distribution cannot represent the case where the probability density of the center of~$\Delta_{K-1}$ is high, depending on the value of~$\hat{\bm{g}}(\bm{x})$, as shown in Fig.~\ref{fig:dir}.
Nevertheless, the samples corresponding to the center of $\Delta_{K-1}$ were given to the probabilistic model, causing $t(\bm{x}, \hat{\bm{\theta}}_{\mathrm{\mu}})$ to be larger than expected, resulting in underconfidence.
On the contrary, the Concrete distribution can represent the case where the probability density of the center of $\Delta_{K-1}$ is high by shifting the temperature parameter $\lambda$ alone, as shown in Fig.~\ref{fig:con}. Therefore, STS can calibrate the confidence accurately.
\section{ESTIMATION OF ALEATORIC AND EPISTEMIC UNCERTAINTIES}
\label{sec:distinction}
As explained in Section~\ref{sec:uncertainty}, the merit of the probabilistic model on $\Delta_{K-1}$ is that it can measure aleatoric uncertainty and epistemic uncertainty separately. A high probability density at the center of $\Delta_{K-1}$ corresponds to aleatoric uncertainty, and a large variance over $\Delta_{K-1}$ corresponds to epistemic uncertainty~\citep{EnD2}. Let $p(\bm{\pi}|\bm{x})$ be a probabilistic model on $\Delta_{K-1}$ with $\bm{x}$-dependent parameters and $\mathbb{E}[\cdot]$ be the expectation of~$p(\bm{\pi}|\bm{x})$. The aleatoric uncertainty is measured by the following expectation of entropy.
\begin{align}
    \mathbb{E}\left[-\sum^K_{k=1}\pi_k\ln \pi_k\right]=-\int\sum^K_{k=1}\pi_k\ln \pi_k\ p(\bm{\pi}|\bm{x}) d\bm{\pi}. \label{eq:ent}
\end{align}
When $p(\bm{\pi}|\bm{x})$ has a high probability density at the center of $\Delta_{K-1}$, Eq.~\eqref{eq:ent} has a large value. In this case, the input $\bm{x}$ is judged to have high aleatoric uncertainty. The epistemic uncertainty is measured by the following differential entropy~\citep{Cover}.
\begin{align}
    \mathbb{E}[-\ln p(\bm{\pi}|\bm{x})]=-\int p(\bm{\pi}|\bm{x})\ln p(\bm{\pi}|\bm{x})d\bm{\pi}. \label{eq:diffent}
\end{align}
When $p(\bm{\pi}|\bm{x})$ has a large variance over $\Delta_{K-1}$, Eq.~\eqref{eq:diffent} has a large value. In this case, the input $\bm{x}$ is judged to have high epistemic uncertainty.

In the case of the proposed probabilistic model shown in Eqs.~\eqref{eq:ind} and~\eqref{eq:con}, all we need to do is set $p(\bm{\pi}|\bm{x})$ to the estimated Concrete distribution $\mathrm{Cn}(\bm{\pi}|\bm{\alpha}(\bm{x}, \hat{\bm{\theta}}_{\mathrm{\alpha}}), \lambda(\bm{x}, \hat{\bm{\theta}}_{\mathrm{\lambda}}))$, in the above metrics.
As in Section~\ref{sec:confidence}, the expectation of the Concrete distribution is approximated as the sample mean of the values sampled by the softmax of the standard Gumbel distribution as follows.
\begin{align}
   \mathbb{E}\left[-\sum^K_{k=1}\pi_k\ln \pi_k\right]&\simeq -\frac{1}{p}\sum^p_{j=1}\sum^K_{k=1}\hat{\pi}^{(j)}_k(\bm{x})\ln \hat{\pi}^{(j)}_k(\bm{x}),\\ 
   \mathbb{E}[-\ln p(\bm{\pi}|\bm{x})]&\simeq - \frac{1}{p}\sum^p_{j=1}\ln \left. \mathrm{Cn}(\bm{\pi}|\bm{\alpha}(\bm{x}, \hat{\bm{\theta}}_{\mathrm{\alpha}}), \lambda(\bm{x}, \hat{\bm{\theta}}_{\mathrm{\lambda}})) \right|_{\bm{\pi}=(\hat{\pi}^{(j)}_1(\bm{x}),\ldots,\hat{\pi}^{(j)}_K(\bm{x}))^\top}, \label{eq:diffent2}
\end{align}
where $\hat{\pi}^{(j)}_k(\bm{x})$ is computed as Eq.~\eqref{eq:pi}.
\begin{table}[t]
    \caption{Pairs of datasets for in-distribution and out-of-distribution (OOD).}
    \label{tab:ood-data}
    \centering
      \begin{tabular}{ccc}                                            \toprule
                                 & in-distribution & out-of-distribution \\\midrule
        Set~1&FashionMNIST~\citep{FMNIST}&    notMNIST~\citep{notMNIST}    \\
        Set~2& CIFAR10~\citep{CIFAR}  &      SVHN~\citep{SVHN}       \\
        Set~3& CIFAR10~\citep{CIFAR}      &   LSUN~\citep{LSUN}  \\\bottomrule
      \end{tabular}
  \end{table}
\begin{table}[t]
    \caption{Area under the curve of precision-recall~(AUPR) and that of receiver operating characteristic~(AUROC) computed by the test datasets for in-distribution and out-of-distribution (OOD).}
    \label{tab:ood-results}
    \centering
      \begin{tabular}{ccccccc}                                            \toprule
           & &  \multirow{2}{*}{pre-trained} & \multirow{2}{*}{PTS} & \multirow{2}{*}{AdaTS}                 & \multicolumn{2}{c}{STS (ours)}   \\\cmidrule{6-7}
         & & & &  & confidence  & differential entropy \\\midrule
        \multirow{2}{*}{Set~1}& AUPR (\%) &  97.80$\pm$0.48 & 97.97$\pm$0.48 & 96.80$\pm$0.94 & 98.02$\pm$0.53 &  \textbf{98.63$\pm$0.46} \\
        & AUROC (\%) &  94.22$\pm$1.28  & 94.30$\pm$1.38 & 89.70$\pm$ 3.30 & \textbf{94.40$\pm$1.56} & \textbf{95.79$\pm$1.43} \\\midrule
        \multirow{2}{*}{Set~2}& AUPR (\%)  & \textbf{98.74$\pm$0.21}  & \textbf{98.84$\pm$0.20} & \textbf{98.89$\pm$0.20} &\textbf{98.81$\pm$0.20} & \textbf{98.93$\pm$0.34} \\
        & AUROC (\%) & \textbf{95.44$\pm$0.51}  & \textbf{95.70$\pm$0.50}  & \textbf{95.74$\pm$0.68} & \textbf{95.51$\pm$0.5}3 & \textbf{95.68$\pm$1.70} \\\midrule
        \multirow{2}{*}{Set~3} & AUPR (\%) & 95.04$\pm$0.15 & 95.39$\pm$0.14 & 94.94$\pm$0.22 & 95.28$\pm$0.17 & \textbf{96.58$\pm$0.21} \\
        & AUROC (\%) & 92.64$\pm$0.17 & 92.93$\pm$0.18 & 92.22$\pm$0.25 & 92.59$\pm$0.23 & \textbf{94.11$\pm$0.37} \\\bottomrule
      \end{tabular}
  \end{table}
\section{EFFECTIVENESS ON OUT-OF-DISTRIBUTION DETECTION}
\label{sec:ood}
In the main paper, Table~\ref{tab:results} shows that STS estimates classification accuracy more precisely than the previous methods. In addition, by computing the differential entropy shown in Eq.~\eqref{eq:diffent2}, STS can achieve higher performance in OOD detection than previous methods, because OOD detection is a task of estimating the epistemic uncertainty that OOD samples have~\citep{uncertainty-survey}.
To show the validity of this claim, we confirmed the OOD detection performance of the pre-trained model and the models calibrated by PTS, AdaTS, and STS.

Table~\ref{tab:ood-data} shows the pairs of datasets for in-distribution and OOD. As shown in Table~\ref{tab:ood-data}, we tried three pairs in our experiments, which are named Set~1, Set~2, and Set~3. Note that the datasets for OOD, notMNIST~\citep{notMNIST}, SVHN~\citep{SVHN}, and LSUN~\citep{LSUN}, were not used in either pre-training, calibration, or tuning of hyperparameters, and these datasets were used for performance evaluation purposes only. The number of samples in the test datasets of notMNIST, SVHN, and LSUN is 18724, 26032, and 10000, respectively. We adopted ResNet18 architecture for the DNN models during pre-training. The training setup was the same as in Section~\ref{sec:ex} and Appendix~\ref{sec:supp}. In other words, we reused the pre-trained model and the calibrated models obtained in the experiments described in the main paper, picking up CIFAR10 and FMNIST. In STS, we also reused the best value of $\beta$ selected as explained in Appendix~\ref{sec:alg}. Therefore, we do not need to train another model for OOD detection. The OOD detection performance was evaluated by the area under the curve of precision-recall~(AUPR) and that of receiver operating characteristic~(AUROC)~\citep{au}. The confidence of the pre-trained model and that of the models calibrated by PTS, AdaTS, and STS were used as OOD detectors. Moreover, differential entropy was also used as an OOD detector in STS. For differential entropy, the same value of $\beta$ was adopted as in the case of confidence.

Table~\ref{tab:ood-results} shows the mean and standard deviation of AUPR and AUROC in the methods. 
As in Table~\ref{tab:results}, bold indicates the best results or results within one standard deviation of the best results.
Although there are no significant differences in Set~2 due to the simplicity of the task, STS shows higher performance than PTS and AdaTS in Set~1 and Set~3. In particular, differential entropy achieved the highest performance, confirming that the differential entropy of a distribution on the probability simplex is effective in OOD detection where a detector that estimates epistemic uncertainty is required. These results also follow up on existing studies claiming the validity of differential entropy on OOD detection~\citep{PriorNet,EnD2}.
\begin{table}[t]
    \caption{Expected Calibration Errors (ECEs) of the test dataset when we performed calibration with different values of $R$ and $S$. The results of the methods except STS are the same as in Table~\ref{tab:results}.}
    \label{tab:RS}
    \centering
      \begin{tabular}{ccccccc}                                            \toprule
          &  \multirow{2}{*}{pre-trained}   & \multirow{2}{*}{PTS} & \multirow{2}{*}{AdaTS} & \multicolumn{3}{c}{STS (ours)}           \\\cmidrule{5-7}
                                 &     &     &  &   $R=10,S=10$ &   $R=20,S=5$  & $R=100,S=1$ \\\midrule
                  FMNIST & 3.86$\pm$0.31 & 2.32$\pm$0.37 & \textbf{1.15$\pm$0.16} & \textbf{1.15$\pm$0.28} &   \textbf{1.01$\pm$0.16} & \textbf{1.04$\pm$0.26} \\
                  CIFAR10 & 2.80$\pm$0.17 & \textbf{1.28$\pm$0.18} & \textbf{1.21$\pm$0.38} & \textbf{1.10$\pm$0.24} & \textbf{1.06$\pm$0.34} & \textbf{1.08$\pm$0.15} \\\bottomrule
      \end{tabular}
  \end{table}

\section{ADDITIONAL EXPERIMENTS WITH DIFFERENT SETUPS}
This section describes calibration experiments with different setups than those described in the previous sections.
\subsection{Sensitivity of Sampling for Multi-Mixup}
\label{sec:RS}
As explained in Appendix~\ref{sec:alg}, in the previous sections we set $R=10$, which is the number of samples in mini-batches $\mathcal{B}^{(k)}$ for each class, and $S=10$, which is the number of times the interpolation is repeated. Since~$R$ and~$S$ are adjustable, we performed additional experiments to investigate the variation in calibration performance as these numbers are adjusted. We used CIFAR10 and FMNIST for the datasets and ResNet18 for the architecture in pre-training. For~$R$ and~$S$, we tried two setups. One is $R=20, S=5$ and the other is $R=100, S=1$. All setups except $R$, $S$, and the number of epochs were not changed from the previous sections. The number of iterations in the optimization was adapted to the experiments in the main paper. Therefore, as we increased~$R$, the number of epochs, which is the number of times each original pre-Multi-Mixup sample is accessed, was increased.

Table~\ref{tab:RS} shows the mean and standard deviation of the ECEs in the methods. There are no significant differences among the different~$R$ and~$S$ settings and the choices of these numbers have little effect on calibration. However, if we focus on the mean, the ECEs become smaller as $R$ is increased. These results suggest the possibility of achieving smaller ECEs.

\subsection{Sensitivity of the Architecture for the Temperature Parameter}
\label{sec:arc}
\begin{figure}[H]
    \centering
      \includegraphics[width=7cm, height=4.3cm]{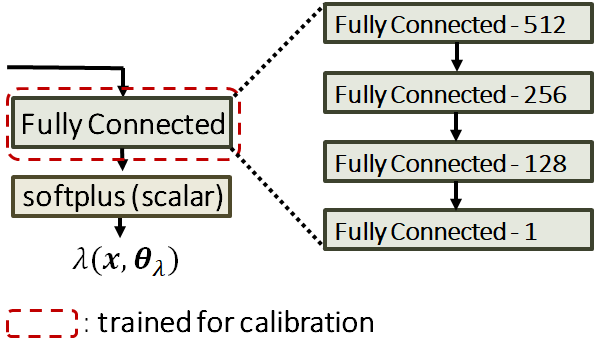}
      \caption{Architecture of the additional branch for the temperature parameter $\lambda(\bm{x}, \bm{\theta}_{\mathrm{\lambda}})$, introduced for additional experiments described in Appendix~\ref{sec:arc}.}
      \label{fig:fully2}
\end{figure}

In the previous sections, we used only the architecture shown in Fig.~\ref{fig:fully} for the temperature parameter $\lambda(\bm{x}, \bm{\theta}_{\mathrm{\lambda}})$. Therefore, this section presents experiments using a different architecture to investigate the sensitivity of the architecture to $\lambda(\bm{x}, \bm{\theta}_{\mathrm{\lambda}})$. As in Appendix~\ref{sec:RS}, we used CIFAR10 and FMNIST for the datasets and ResNet18 for the architecture in pre-training. Figure~\ref{fig:fully2} shows a new architecture introduced for additional experiments. The new architecture has fewer layers and narrower widths than the architecture in Fig.~\ref{fig:fully}. All setups except architectures were not changed from the previous sections. We adopted the $R=10,S=10$ setting.

Table~\ref{tab:arc} shows the mean and standard deviation of the ECEs in the methods. There are no significant differences among the different architectures and the choices of the architectures for $\lambda(\bm{x}, \bm{\theta}_{\mathrm{\lambda}})$ have little effect on calibration. However, if we focus on the mean, the ECEs get larger as the architecture gets smaller. This suggests that there is a risk that the calibration performance will deteriorate as the architecture gets smaller, and we should make sure that the size of the architecture is sufficient for $\lambda(\bm{x}, \bm{\theta}_{\mathrm{\lambda}})$ when we perform the calibration using STS.
\newpage

\begin{table}[H]
    \caption{Expected Calibration Errors (ECEs) of the test dataset when we performed calibration with different architectures for the temperature parameter $\lambda(\bm{x}, \bm{\theta}_{\mathrm{\lambda}})$. The results of the methods except STS are the same as in Table~\ref{tab:results}. The columns indicated by ``Figure~\ref{fig:fully}'' and ``Figure~\ref{fig:fully2}'' show the results with the different architectures.}
    \label{tab:arc}
    \centering
      \begin{tabular}{cccccc}                                            \toprule
        \  &  \multirow{2}{*}{pre-trained}   & \multirow{2}{*}{PTS} & \multirow{2}{*}{AdaTS} & \multicolumn{2}{c}{STS (ours)}           \\\cmidrule{5-6}
                                 &     &     &  &   Figure~\ref{fig:fully} & Figure~\ref{fig:fully2}  \\\midrule
                  FMNIST & 3.86$\pm$0.31 & 2.32$\pm$0.37 & \textbf{1.15$\pm$0.16} & \textbf{1.15$\pm$0.28} & \textbf{1.30$\pm$0.18}  \\
                  CIFAR10 & 2.80$\pm$0.17 & \textbf{1.28$\pm$0.18} & \textbf{1.21$\pm$0.38} & \textbf{1.10$\pm$0.24} & \textbf{1.11$\pm$0.16}  \\\bottomrule
      \end{tabular}
  \end{table}
\vfill


\begin{thebibliography}{}

\bibitem[Abhishek et~al., 2022]{zeta-mixup}
Abhishek, K., Brown, C.~J., and Hamarneh, G. (2022).
\newblock Multi-sample $\zeta$-mixup: Richer, more realistic synthetic samples
  from a $p$-series interpolant.
\newblock {\em arXiv preprint arXiv: 2204.03323}.

\bibitem[Balanya et~al., 2022]{AdaptiveTemp2}
Balanya, S.~A., Maro^^c3^^b1as, J., and Ramos, D. (2022).
\newblock Adaptive temperature scaling for robust calibration of deep neural
  networks.
\newblock {\em arXiv preprint arXiv: 2208.00461}.

\bibitem[Bottou, 2010]{SGD}
Bottou, L. (2010).
\newblock Large-scale machine learning with stochastic gradient descent.
\newblock In {\em International Conference on Computational Statistics}, pages
  177--186. Physica-Verlag HD.

\bibitem[Boureau et~al., 2010]{pooling}
Boureau, Y., Ponce, J., and LeCun, Y. (2010).
\newblock A theoretical analysis of feature pooling in visual recognition.
\newblock In {\em International Conference on Machine Learning}, page
  111^^e2^^80^^93118. Omnipress.

\bibitem[Bulatov, 2011]{notMNIST}
Bulatov, Y. (2011).
\newblock not{MNIST} dataset.
\newblock Google (Books/OCR), Tech. Rep.
  \url{https://yaroslavvb.blogspot.com/2011/09/notmnist-dataset.html}, Accessed
  on February 1, 2024.

\bibitem[Coates et~al., 2011]{STL10}
Coates, A., Ng, A., and Lee, H. (2011).
\newblock An analysis of single-layer networks in unsupervised feature
  learning.
\newblock In {\em International Conference on Artificial Intelligence and
  Statistics}, volume~15, pages 215--223. PMLR.

\bibitem[Cover and Thomas, 2006]{Cover}
Cover, T.~M. and Thomas, J.~A. (2006).
\newblock {\em Elements of Information Theory}.
\newblock Wiley-Interscience.

\bibitem[Filho et~al., 2021]{calibration-survay}
Filho, T.~S., Song, H., Perello-Nieto, M., Santos-Rodriguez, R., Kull, M., and
  Flach, P. (2021).
\newblock Classifier calibration: A survey on how to assess and improve
  predicted class probabilities.
\newblock {\em arXiv preprint arXiv: 2112.10327}.

\bibitem[Gawlikowski et~al., 2021]{uncertainty-survey}
Gawlikowski, J., Tassi, C. R.~N., Ali, M., Lee, J., Humt, M., Feng, J., Kruspe,
  A., Triebel, R., Jung, P., Roscher, R., Shahzad, M., Yang, W., Bamler, R.,
  and Zhu, X.~X. (2021).
\newblock A survey of uncertainty in deep neural networks.
\newblock {\em arXiv preprint arXiv: 2107.03342}.

\bibitem[Glorot et~al., 2011]{softplus}
Glorot, X., Bordes, A., and Bengio, Y. (2011).
\newblock Deep sparse rectifier neural networks.
\newblock In {\em International Conference on Artificial Intelligence and
  Statistics}, volume~15, pages 315--323. PMLR.

\bibitem[Gumbel, 1941]{Gumbel}
Gumbel, E.~J. (1941).
\newblock The return period of flood flows.
\newblock {\em The Annals of Mathematical Statistics}, 12(2):163--190.

\bibitem[Guo et~al., 2017]{calibration}
Guo, C., Pleiss, G., Sun, Y., and Weinberger, K.~Q. (2017).
\newblock On calibration of modern neural networks.
\newblock In {\em International Conference on Machine Learning}, volume~70,
  pages 1321--1330. PMLR.

\bibitem[Guo et~al., 2019]{AdaMixUp}
Guo, H., Mao, Y., and Zhang, R. (2019).
\newblock Mixup as locally linear out-of-manifold regularization.
\newblock {\em AAAI Conference on Artificial Intelligence}, 33(01):3714--3722.

\bibitem[H^^c3^^bcllermeier and Waegeman, 2021]{uncertainty}
H^^c3^^bcllermeier, E. and Waegeman, W. (2021).
\newblock Aleatoric and epistemic uncertainty in machine learning: an
  introduction to concepts and methods.
\newblock {\em Machine Learning}, 110:457--506.

\bibitem[He et~al., 2016]{ResNet}
He, K., Zhang, X., Ren, S., and Sun, J. (2016).
\newblock Deep residual learning for image recognition.
\newblock In {\em IEEE Conference on Computer Vision and Pattern Recognition}.

\bibitem[Huang et~al., 2017]{DenseNet}
Huang, G., Liu, Z., Maaten, L., and Weinberger, K.~Q. (2017).
\newblock Densely connected convolutional networks.
\newblock In {\em Conference on Computer Vision and Pattern Recognition}.

\bibitem[Inoue, 2018]{PairingSamples}
Inoue, H. (2018).
\newblock Data augmentation by pairing samples for images classification.
\newblock {\em arXiv preprint arXiv: 1801.02929}.

\bibitem[Jang et~al., 2017]{gumbelsoft}
Jang, E., Gu, S., and Poole, B. (2017).
\newblock Categorical reparameterization with gumbel-softmax.
\newblock In {\em International Conference on Learning Representations}.

\bibitem[Joy, 2022]{AdaTS-code}
Joy, T. (2022).
\newblock Adaptive temperature scaling [{S}ource code].
\newblock \url{https://github.com/thwjoy/adats}, Accessed on July 21, 2023.

\bibitem[Joy et~al., 2023]{AdaptiveTemp}
Joy, T., Pinto, F., Lim, S., Torr, P.~H., and Dokania, P.~K. (2023).
\newblock Sample-dependent adaptive temperature scaling for improved
  calibration.
\newblock {\em AAAI Conference on Artificial Intelligence},
  37(12):14919--14926.

\bibitem[Kingma and Ba, 2014]{Adam}
Kingma, D.~P. and Ba, J. (2014).
\newblock Adam: A method for stochastic optimization.
\newblock {\em arXiv preprint arXiv: 1412.6980}.

\bibitem[Kristiadi et~al., 2020]{overconfidence}
Kristiadi, A., Hein, M., and Hennig, P. (2020).
\newblock Being bayesian, even just a bit, fixes overconfidence in {R}e{LU}
  networks.
\newblock In {\em International Conference on Machine Learning}, volume 119,
  pages 5436--5446. PMLR.

\bibitem[Krizhevsky and Hinton, 2009]{CIFAR}
Krizhevsky, A. and Hinton, G. (2009).
\newblock Learning multiple layers of features from tiny images.
\newblock Technical report, University of Toronto.

\bibitem[Kuleshov and Deshpande, 2022]{DensityEstimation}
Kuleshov, V. and Deshpande, S. (2022).
\newblock Calibrated and sharp uncertainties in deep learning via density
  estimation.
\newblock In {\em International Conference on Machine Learning}, volume 162,
  pages 11683--11693. PMLR.

\bibitem[Lecun et~al., 1998]{LeNet}
Lecun, Y., Bottou, L., Bengio, Y., and Haffner, P. (1998).
\newblock Gradient-based learning applied to document recognition.
\newblock {\em Proceedings of the IEEE}, 86(11):2278--2324.

\bibitem[Lindqvist et~al., 2020]{generalEnD2}
Lindqvist, J., Olmin, A., Lindsten, F., and Svensson, L. (2020).
\newblock A general framework for ensemble distribution distillation.
\newblock In {\em International Workshop on Machine Learning for Signal
  Processing}, pages 1--6.

\bibitem[Loshchilov and Hutter, 2017]{CosineAnnealing}
Loshchilov, I. and Hutter, F. (2017).
\newblock {SGDR}: Stochastic gradient descent with warm restarts.
\newblock In {\em International Conference on Learning Representations}.

\bibitem[Maddison et~al., 2017]{concrete}
Maddison, C.~J., Mnih, A., and Teh, Y.~W. (2017).
\newblock The concrete distribution: A continuous relaxation of discrete random
  variables.
\newblock In {\em International Conference on Learning Representations}.

\bibitem[Malinin and Gales, 2018]{PriorNet}
Malinin, A. and Gales, M. (2018).
\newblock Predictive uncertainty estimation via prior networks.
\newblock In {\em Advances in Neural Information Processing Systems},
  volume~31. Curran Associates, Inc.

\bibitem[Malinin et~al., 2020]{EnD2}
Malinin, A., Mlodozeniec, B., and Gales, M. (2020).
\newblock Ensemble distribution distillation.
\newblock In {\em International Conference on Learning Representations}.

\bibitem[Netzer et~al., 2011]{SVHN}
Netzer, Y., Wang, T., Coates, A., Bissacco, A., Wu, B., and Ng, A.~Y. (2011).
\newblock Reading digits in natural images with unsupervised feature learning.
\newblock In {\em NIPS Workshop on Deep Learning and Unsupervised Feature
  Learning}.

\bibitem[Paszke et~al., 2019]{PyTorch}
Paszke, A., Gross, S., Massa, F., Lerer, A., Bradbury, J., Chanan, G., Killeen,
  T., Lin, Z., Gimelshein, N., Antiga, L., Desmaison, A., Kopf, A., Yang, E.,
  DeVito, Z., Raison, M., Tejani, A., Chilamkurthy, S., Steiner, B., Fang, L.,
  Bai, J., and Chintala, S. (2019).
\newblock {PyTorch}: An imperative style, high-performance deep learning
  library.
\newblock In {\em Advances in Neural Information Processing Systems},
  volume~32. Curran Associates, Inc.

\bibitem[Polak, 1997]{gradient-descent}
Polak, E. (1997).
\newblock {\em Optimization : Algorithms and Consistent Approximations}.
\newblock Springer-Verlag.

\bibitem[Powers, 2020]{au}
Powers, D. M.~W. (2020).
\newblock Evaluation: from precision, recall and f-measure to roc,
  informedness, markedness and correlation.
\newblock {\em arXiv preprint arXiv: 2010.16061}.

\bibitem[Ramalho and Miranda, 2020]{DS}
Ramalho, T. and Miranda, M. (2020).
\newblock Density estimation in representation space to predict model
  uncertainty.
\newblock In {\em Engineering Dependable and Secure Machine Learning Systems},
  pages 84--96. Springer International Publishing.

\bibitem[Ryabinin et~al., 2021]{ScalingEnD2}
Ryabinin, M., Malinin, A., and Gales, M. (2021).
\newblock Scaling ensemble distribution distillation to many classes with proxy
  targets.
\newblock In {\em Advances in Neural Information Processing Systems},
  volume~34, pages 6023--6035. Curran Associates, Inc.

\bibitem[Sensoy et~al., 2018]{EDL}
Sensoy, M., Kaplan, L., and Kandemir, M. (2018).
\newblock Evidential deep learning to quantify classification uncertainty.
\newblock In {\em Advances in Neural Information Processing Systems},
  volume~31. Curran Associates, Inc.

\bibitem[Simonyan and Zisserman, 2014]{VGG}
Simonyan, K. and Zisserman, A. (2014).
\newblock Very deep convolutional networks for large-scale image recognition.
\newblock {\em arXiv preprint arXiv: 1409.1556}.

\bibitem[Tan et~al., 2018]{transfer-learning}
Tan, C., Sun, F., Kong, T., Zhang, W., Yang, C., and Liu, C. (2018).
\newblock A survey on deep transfer learning.
\newblock In {\em International Conference on Artificial Neural Networks},
  pages 270--279. Springer International Publishing.

\bibitem[Thulasidasan et~al., 2019]{mixup-calibration}
Thulasidasan, S., Chennupati, G., Bilmes, J.~A., Bhattacharya, T., and
  Michalak, S. (2019).
\newblock On mixup training: Improved calibration and predictive uncertainty
  for deep neural networks.
\newblock In {\em Advances in Neural Information Processing Systems},
  volume~32. Curran Associates, Inc.

\bibitem[Tomani et~al., 2022]{PTS}
Tomani, C., Cremers, D., and Buettner, F. (2022).
\newblock Parameterized temperature scaling for boosting the expressive power
  in post-hoc uncertainty calibration.
\newblock In {\em European Conference on Computer Vision}, pages 555--569.
  Springer Nature Switzerland.

\bibitem[Verma et~al., 2019]{ManifoldMixup}
Verma, V., Lamb, A., Beckham, C., Najafi, A., Mitliagkas, I., Lopez-Paz, D.,
  and Bengio, Y. (2019).
\newblock Manifold mixup: Better representations by interpolating hidden
  states.
\newblock In {\em International Conference on Machine Learning}, volume~97,
  pages 6438--6447. PMLR.

\bibitem[Wenger et~al., 2020]{non-parametric}
Wenger, J., Kjellstr\"om, H., and Triebel, R. (2020).
\newblock Non-parametric calibration for classification.
\newblock In {\em International Conference on Artificial Intelligence and
  Statistics}, volume 108, pages 178--190. PMLR.

\bibitem[Wilson and Izmailov, 2020]{generalization}
Wilson, A.~G. and Izmailov, P. (2020).
\newblock Bayesian deep learning and a probabilistic perspective of
  generalization.
\newblock In {\em Advances in Neural Information Processing Systems},
  volume~33, pages 4697--4708. Curran Associates, Inc.

\bibitem[Xiao et~al., 2017]{FMNIST}
Xiao, H., Rasul, K., and Vollgraf, R. (2017).
\newblock Fashion-{MNIST}: A novel image dataset for benchmarking machine
  learning algorithms.
\newblock {\em arXiv preprint arXiv: 1708.07747}.
\newblock The MIT License (MIT) Copyright \copyright\ 2017 Zalando SE,
  \url{https://tech.zalando.com}.

\bibitem[Yang et~al., 2021]{ood-survay}
Yang, J., Zhou, K., Li, Y., and Liu, Z. (2021).
\newblock Generalized out-of-distribution detection: A survey.
\newblock {\em arXiv preprint arXiv: 2110.11334}.

\bibitem[Yu et~al., 2015]{LSUN}
Yu, F., Seff, A., Zhang, Y., Song, S., Funkhouser, T., and Xiao, J. (2015).
\newblock Lsun: Construction of a large-scale image dataset using deep learning
  with humans in the loop.
\newblock {\em arXiv preprint arXiv: 1506.03365}.

\bibitem[Zhang et~al., 2018]{mixup}
Zhang, H., Cisse, M., Dauphin, Y.~N., and Lopez-Paz, D. (2018).
\newblock mixup: Beyond empirical risk minimization.
\newblock In {\em International Conference on Learning Representations}.

\bibitem[Zhang et~al., 2020]{ETS}
Zhang, J., Kailkhura, B., and Han, T.~Y. (2020).
\newblock Mix-n-{M}atch : Ensemble and compositional methods for uncertainty
  calibration in deep learning.
\newblock In {\em International Conference on Machine Learning}, volume 119,
  pages 11117--11128. PMLR.

\bibitem[Zou et~al., 2019]{self-training}
Zou, Y., Yu, Z., Liu, X., Kumar, B.~V., and Wang, J. (2019).
\newblock Confidence regularized self-training.
\newblock In {\em IEEE/CVF International Conference on Computer Vision}.

\end{thebibliography}
\end{document}